\documentclass[10pt]{article}
\usepackage[utf8]{inputenc}
\usepackage[T1]{fontenc}
\usepackage{microtype}
\usepackage{amsmath}
\usepackage{amssymb}
\usepackage{graphicx}
\usepackage{xcolor}
\usepackage{booktabs}

\usepackage{colortbl}
\usepackage{pifont}
\providecommand{\cmark}{{\color{black}\ding{51}}}
\providecommand{\pmark}{{\color{gray}(\ding{51})}}
\providecommand{\xmark}{{\color{gray!60}\ding{55}}}
\usepackage{longtable}
\usepackage{array}
\usepackage{adjustbox}
\usepackage{enumitem}
\usepackage{listings}
\usepackage[numbers,sort&compress]{natbib}
\usepackage{caption}
\usepackage[colorlinks=true,linkcolor=blue!55!black,citecolor=blue!55!black,urlcolor=blue!55!black,bookmarksnumbered=true]{hyperref}

\newcommand{\mdw}[1]{\dimexpr #1\linewidth-2\tabcolsep\relax}
\setlist[itemize]{leftmargin=1.6em,itemsep=0.15em,topsep=0.3em,parsep=0.1em}
\setlist[enumerate]{leftmargin=1.9em,itemsep=0.15em,topsep=0.3em,parsep=0.1em}
\definecolor{mdrulegray}{gray}{0}
\definecolor{mdarrowgray}{gray}{0}
\newif\ifshowtodos
\showtodostrue
\DeclareRobustCommand{\todo}[2][TODO]{\ifshowtodos\begingroup\def\_{\textunderscore\allowbreak{}}%
  \textcolor{red}{[#1\if\relax\detokenize{#2}\relax\else: #2\fi]}\endgroup\fi}

\newcounter{mdbox}

\newsavebox{\mdboxbox}
\newenvironment{mdbox}[2]
  {\par\medskip\begingroup\renewcommand{\themdbox}{#1}\refstepcounter{mdbox}\label{#2}%
   \begin{lrbox}{\mdboxbox}\begin{minipage}{\dimexpr\linewidth-2\fboxsep-2\fboxrule\relax}%
   \small\setlength{\parindent}{0pt}\setlength{\parskip}{0.4em}}
  {\end{minipage}\end{lrbox}\noindent\fbox{\usebox{\mdboxbox}}\endgroup\par\medskip}
\makeatletter
\newif\if@mdappendix
\@mdappendixfalse
\let\mdorig@appendix\appendix
\renewcommand{\appendix}{\mdorig@appendix\@mdappendixtrue}
\newcommand{\mdsec@section}{}
\renewcommand{\@seccntformat}[1]{%
  \if@mdappendix\ifcsname mdsec@#1\endcsname Appendix~\fi\fi
  \csname the#1\endcsname
  \if@mdappendix\ifcsname mdsec@#1\endcsname .\fi\fi
  \quad}
\makeatother
\newif\ifmdbibprinted
\mdbibprintedfalse
\newif\ifmdcitations
\mdcitationsfalse
\newcommand{\mdBibStyle}{plainnat}
\newcommand{\mdBibFiles}{refs,refs_bench}
\newcommand{\mdBibliography}{\ifmdcitations\ifmdbibprinted\else\mdbibprintedtrue
  \bibliographystyle{\mdBibStyle}\bibliography{\mdBibFiles}\fi\fi}

\usepackage{startlux-bench}

\begin{document}
\title{DecisionPivot: A Diagnostic Benchmark for Decision Models in LLM Systems}
% Title block and resource row as in the technical report (startlux-bench.sty). One shared affiliation, so the names
% carry no affiliation numbers; * marks the corresponding authors (footnote in main.tex). Two lines of names, then
% the affiliation.
\author{Xialie Zhuang\quad\ Mingcong Lei\quad\ Zunhai Su\quad\ Ri Liu\quad\ Jianjin Li\\
Peixian Ma\quad\ Jia Pan\quad Ngai Wong\quad\ Guanhua Chen\textsuperscript{*}\quad\ Chuan-Wei Kuo\textsuperscript{*}\\[6pt]
{\normalsize StartLux}}
\date{October 2026}
\hypersetup{pdftitle={DecisionPivot: A Diagnostic Benchmark for Decision Models in LLM Systems},pdfauthor={Xialie Zhuang, Mingcong Lei, Zunhai Su, Ri Liu, Jianjin Li, Peixian Ma, Jia Pan,  Guanhua Chen,Chuan-Wei Kuo}}
\mdcitationstrue
\begin{mdresources}
\mdresource{Data}{hf.pdf}{\href{https://huggingface.co/datasets/startlux-models/DecisionPivot}{Hugging Face}}
\mdresource{Code}{github.pdf}{\href{https://github.com/StartLuxLabs/DecisionPivot}{GitHub}}
\end{mdresources}
\maketitle
\begingroup
\renewcommand{\thefootnote}{*}
\footnotetext{Corresponding authors.}
\endgroup
\begin{abstract}
Decision models support routing, guardrails, and output verification in LLM systems. However, item accuracy alone cannot establish whether models make correct decisions across controlled variations or assign reliable confidence to their predictions. We introduce DecisionPivot, a diagnostic benchmark comprising 48 tasks and 40,033 items across seven decision roles. Its core track contains 7,800 items from 13 generated task families, organized into contrast groups with computed answers verified by independently implemented solvers. The complementary reference track contains 32,233 items from 35 public test sets, preserving their original evaluation metrics. We evaluate 40 decision models and 14 general models using documented, model-specific protocols, assessing item and contrast-group accuracy, probability ordering, calibration, and empirical selective coverage. Our results reveal substantial differences obscured by aggregate scores: two 27B models separated by only 2.6 points on the reference track achieve contrast-group accuracies of 70.1\% and 44.2\%. Across 2,400 binary contrast pairs, Perplexity Decider correctly ranks the positive member above the negative member in 92.3\% of pairs but answers both members correctly in only 74.1\%, exposing a gap between probability ordering and decision correctness. Models with similar item accuracy also differ in the reliability of their high-confidence predictions. On a separate 1,296-item subset, GLM-5.3 achieves 93.5\% contrast-group accuracy at a higher reasoning setting, compared with 81.5\% at its lowest effort. By linking errors to pivotal details and difficulty levels, DecisionPivot provides fine-grained diagnostics of decision failures that aggregate evaluation alone cannot reveal.
\end{abstract}

\FloatBarrier
\section{Introduction}\label{sec:1}

LLM systems make decisions throughout the processing of a request: which model should handle it, whether a tool call is permitted, whether a generated output is correct, and whether an agent has achieved its goal.
A \emph{decision model} takes a state, a typed question, and a finite set of options as input and assigns a probability to each option.
Many such models produce these predictions in a single forward pass, enabling systems to act directly or defer uncertain cases for further reasoning or review.
Reliable evaluation must therefore assess not only decision correctness but also sensitivity to decisive evidence and the reliability of predicted confidence.

Existing evaluations adapt public test sets into decision tasks and assess calibration and selective prediction \citep{deusser2026evaluating}.
While these evaluations provide valuable comparisons, aggregate item accuracy does not reveal whether a model correctly handles all members of a controlled contrast group.
Moreover, superficial cues can inflate performance: a content-blind rule achieves 73.0\% accuracy on \texttt{llmbar}, compared with a 50\% chance baseline (Table~\ref{tab:C2}).
Potential training overlap further complicates interpretation, as public documentation may identify training data sources without specifying which evaluation items were included.
These limitations motivate combining established public benchmarks with controlled, independently verifiable decision cases.

We introduce \emph{DecisionPivot}, a diagnostic benchmark spanning seven decision roles and comprising two complementary tracks.
The \emph{core track} contains 7,800 items from 13 generated task families, organized into contrast groups.
In most families, modifying a \emph{pivotal detail} changes the correct decision while preserving the surrounding context, enabling controlled evaluation across related cases.
Independently implemented solvers verify every core answer.
The \emph{reference track} contains 32,233 items from 35 public test sets, preserving their original evaluation metrics and available training-overlap information.
Together, these tracks support controlled diagnosis while maintaining comparability with established evaluations.

DecisionPivot evaluates three complementary aspects of decision quality: whether individual predictions are correct, whether all members of a contrast group are answered correctly, and whether predicted confidence reliably reflects correctness.
We measure group-level correctness using \emph{pair accuracy}, which requires every member of a contrast group to be correct, including groups containing more than two items.
We further assess probability ordering, calibration, and selective coverage to distinguish correct relative rankings from correct decisions and evaluate confidence reliability.

\begin{figure}[!tbp]
\renewcommand{\thefigure}{1}
\centering
\adjincludegraphics[max width=\linewidth,max height=0.5\textheight]{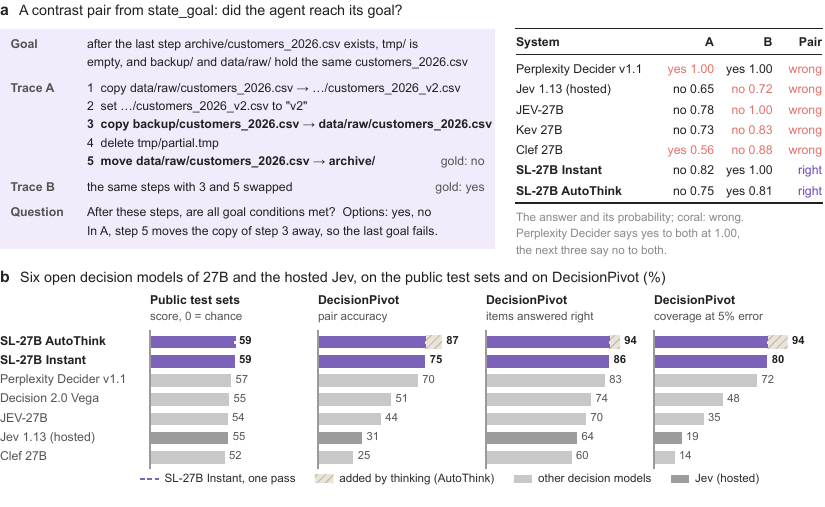}
\caption{\textbf{Controlled contrasts reveal errors obscured by aggregate scores.} (a) A level-1 \texttt{state\_goal} contrast pair differing only in the order of steps 3 and 5. In A, step 5 removes the file copied in step 3, violating a final goal condition; in B, the goal is satisfied. Predictions and probabilities are shown, with incorrect predictions highlighted in red. Only the two SL-27B variants among the displayed models correctly answer both members. (b) Comparison of six open 27B decision models and the hosted Jev reference across four metrics: reference score (0 = chance), core pair accuracy, core item accuracy, and empirical core coverage at a 5\% error rate. The displayed models span approximately 6.9 reference-score points and 62 percentage points in core pair accuracy. Results for all 40 decision models are reported in Table~\ref{tab:4}.}\label{fig:1}
\end{figure}

We evaluate 40 decision models and 11 open general models on both tracks, plus three API chat models on the core track, using documented, model-specific protocols.
Our results reveal substantial differences that aggregate scores alone do not capture (Figure~\ref{fig:1}).
Perplexity Decider and JEV-27B achieve similar reference scores but differ markedly in core pair accuracy.
Kev 27B and JEV-27B exhibit nearly identical item accuracy yet differ in the frequency and correctness of high-confidence predictions.
Binary contrast analysis further reveals that correct probability ordering does not necessarily translate into correct decisions.
Comparisons across reasoning settings show improvements on selected cases, highlighting the value of evaluating decision behavior beyond aggregate accuracy.

We make three contributions:
\begin{itemize}
    \item \textbf{A controlled benchmark with independently verified answers.}
    We introduce DecisionPivot, comprising 13 generated task families across seven decision roles and 35 complementary public test sets.
    Controlled contrast groups, annotations of pivotal variations, and independent answer verification enable fine-grained diagnosis of decision failures (Section~\ref{sec:3}).

    \item \textbf{A unified framework for diagnosing decision reliability.}
    We jointly evaluate item and contrast-group correctness, relative probability ordering, calibration, and selective coverage.
    These complementary metrics distinguish failures in decision correctness, probability ranking, and confidence reliability that aggregate accuracy alone cannot fully characterize (Section~\ref{sec:3.4}).

    \item \textbf{Empirical insights into decision-model failures.}
    Evaluating 40 decision models and 14 general models, we reveal substantial differences in controlled-contrast performance, confidence reliability, and sensitivity to task difficulty, reasoning settings, and input presentation.
    Fresh samples from seven generators further assess the stability of model rankings across sampled cases (Sections~\ref{sec:4} and \ref{sec:5}).
\end{itemize}

\FloatBarrier
\section{Related work}\label{sec:2}

\textbf{Decision-model and task-specific evaluation.}
\citet{deusser2026evaluating} evaluate a hosted decision model on 37 public datasets, examining calibration, selective prediction, and option-order sensitivity, with bootstrap confidence intervals.
The \href{https://github.com/apolinario/decision-index}{Decision Index} aggregates public benchmarks into a chance-corrected index, while \href{https://github.com/fstandhartinger/jevbench}{JevBench} evaluates calibration, speed, and cost on its own items.
These evaluations provide useful comparisons, while DecisionPivot complements them with programmatically verified cases and group-level tests of controlled outcome changes (Table~\ref{tab:1}).
Task-specific benchmarks address routing \citep{hu2024routerbench}, guardrails \citep{bassani2024guardbench}, reward modeling \citep{lambert2024rewardbench}, response judging \citep{tan2024judgebench}, abstention \citep{kirichenko2025abstentionbench}, legal reasoning \citep{guha2023legalbench}, and retrieval \citep{thakur2021beir}.
Our reference track adapts selected public tasks to a common decision interface while preserving their original scoring criteria.

\textbf{Generated cases and controlled contrasts.}
Programmatically constructed tasks such as bAbI \citep{weston2015towards} provide answers known by construction, while LiveBench \citep{white2024livebench} refreshes evaluation questions to limit contamination.
GSM-Symbolic \citep{mirzadeh2024gsm} examines performance variation across instances of mathematical templates.
Contrast sets \citep{gardner2020evaluating} and counterfactually augmented data \citep{kaushik2019learning} study model behavior under label-changing modifications.
Partial-input baselines \citep{gururangan2018annotation} and heuristic diagnostics \citep{mccoy2019right} reveal shortcuts that can inflate conventional evaluation scores.
DecisionPivot brings these complementary ideas together in typed decision tasks, combining controlled contrasts, independently verified answers, group-level evaluation, and annotations of pivotal variations.

\textbf{Confidence and behavioral robustness.}
Calibration analysis \citep{guo2017calibration} and selective classification \citep{geifman2017selective} provide established methods for assessing confidence reliability and abstention.
We combine these measures with contrast-group correctness and probability ordering to characterize complementary aspects of decision reliability.
Our input-sensitivity analyses build on studies of option-order effects \citep{zheng2023large}, irrelevant context \citep{shi2023large}, and behavioral testing \citep{ribeiro2020beyond}.
These analyses examine whether stable aggregate scores also imply stable individual predictions.

\begin{table}[!tbp]
\renewcommand{\thetable}{1}
\centering
\caption{\textbf{Comparison of evaluation features across related benchmarks.} \cmark{} supported, \pmark{} partially supported, \xmark{} not documented in the reviewed sources. Typed decisions use a fixed set of options and return a probability for each option. Probability scoring includes calibration or other probability-based metrics. Selective decisions assess abstention or coverage at a specified error rate. Contamination control denotes measures to limit prior exposure, such as private, refreshed, or generated items; it does not establish the absence of all training overlap. Features may apply to only part of a benchmark; DecisionPivot's computed answers and controlled contrasts refer to its core track.}\label{tab:1}
{\footnotesize\setlength{\tabcolsep}{6pt}\renewcommand{\arraystretch}{1.15}
\noindent\begin{adjustbox}{max width=\linewidth}
\begin{tabular}{l*{8}{c}}
\toprule
\textbf{benchmark} & \rotatebox{90}{\textbf{typed decisions}} & \rotatebox{90}{\textbf{computed answers}} & \rotatebox{90}{\textbf{contrast groups}} & \rotatebox{90}{\textbf{difficulty levels}} & \rotatebox{90}{\textbf{probability scored}} & \rotatebox{90}{\textbf{selective decisions}} & \rotatebox{90}{\textbf{contamination control}} & \rotatebox{90}{\textbf{input variants}} \\
\midrule
\mdbodyrow \citet{deusser2026evaluating} & \cmark & \xmark & \xmark & \xmark & \cmark & \cmark & \pmark & \pmark \\
\mdbodyrow Decision Index 0.3 & \cmark & \xmark & \xmark & \xmark & \cmark & \xmark & \cmark & \xmark \\
\mdbodyrow JevBench 1.6 & \cmark & \xmark & \xmark & \cmark & \cmark & \pmark & \cmark & \xmark \\
\mdbodyrow AbstentionBench \citep{kirichenko2025abstentionbench} & \xmark & \xmark & \xmark & \xmark & \xmark & \cmark & \xmark & \xmark \\
\mdbodyrow Contrast sets \citep{gardner2020evaluating} & \xmark & \xmark & \cmark & \xmark & \xmark & \xmark & \xmark & \xmark \\
\mdbodyrow GSM-Symbolic \citep{mirzadeh2024gsm} & \xmark & \cmark & \xmark & \cmark & \xmark & \xmark & \pmark & \cmark \\
\mdbodyrow LiveBench \citep{white2024livebench} & \xmark & \pmark & \xmark & \xmark & \xmark & \xmark & \cmark & \xmark \\
\mdourrow \textbf{DecisionPivot (this work)} & \cmark & \cmark & \cmark & \cmark & \cmark & \cmark & \cmark & \cmark \\
\bottomrule
\end{tabular}
\end{adjustbox}\par}
\end{table}%
\FloatBarrier
\section{DecisionPivot}\label{sec:3}

DecisionPivot comprises two complementary tracks under a common typed decision interface: a generated core track for controlled diagnosis and a public reference track for comparison on established tasks.
Both tracks span seven functional decision roles (Figure~\ref{fig:2}).

\subsection{Items and decision layers}\label{sec:3.1}

Each item consists of a state \(s\), a typed question \(q\), a finite option set \(O\), and an answer label.
Core items have a gold option \(y\in O\), while reference tasks may also provide success-rate labels or ranking judgments evaluated using their original metrics.
A model produces a probability distribution \(p(o\mid s,q,O)\) over the options, with the predicted answer given by \(\hat y=\arg\max_{o\in O}p(o\mid s,q,O)\).
Core items additionally record their contrast-group membership, difficulty level, and, where available, the type of pivotal variation.

\begin{figure}[!tbp]
\renewcommand{\thefigure}{2}
\centering
\adjincludegraphics[max width=\linewidth,max height=0.5\textheight]{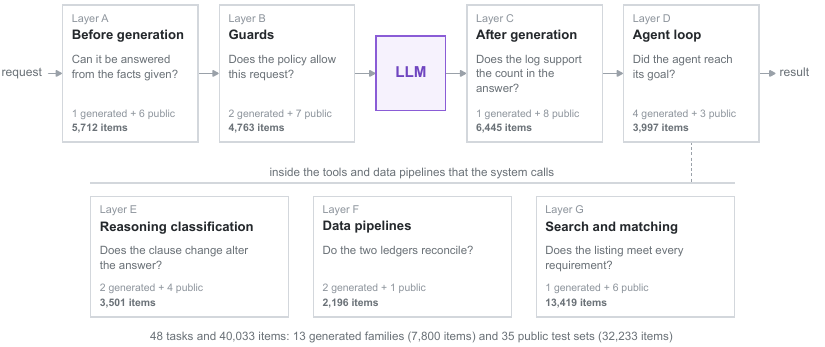}
\caption{\textbf{DecisionPivot covers 48 tasks across seven decision roles.} Layers A--D span the request-to-result workflow, while layers E--G cover functions within tools and data pipelines. Each box illustrates a representative decision question and reports the numbers of generated task families, public test sets, and items.}\label{fig:2}
\end{figure}

We organize tasks into seven functional roles within LLM systems (Figure~\ref{fig:2}).
Layers A--D span the request-to-result workflow: answerability and routing (A), policy-based guardrails (B), output verification (C), and tool validation or goal checking in agent loops (D).
Layers E--G cover reasoning-based classification (E), record integrity and reconciliation (F), and constraint-based search and matching (G) within tools and data pipelines.
These layers characterize functional roles rather than prescribing a specific system architecture.

\subsection{Core track}\label{sec:3.2.1}

\begin{figure}[!htbp]
\renewcommand{\thefigure}{3}
\centering
\adjincludegraphics[max width=\linewidth,max height=0.5\textheight]{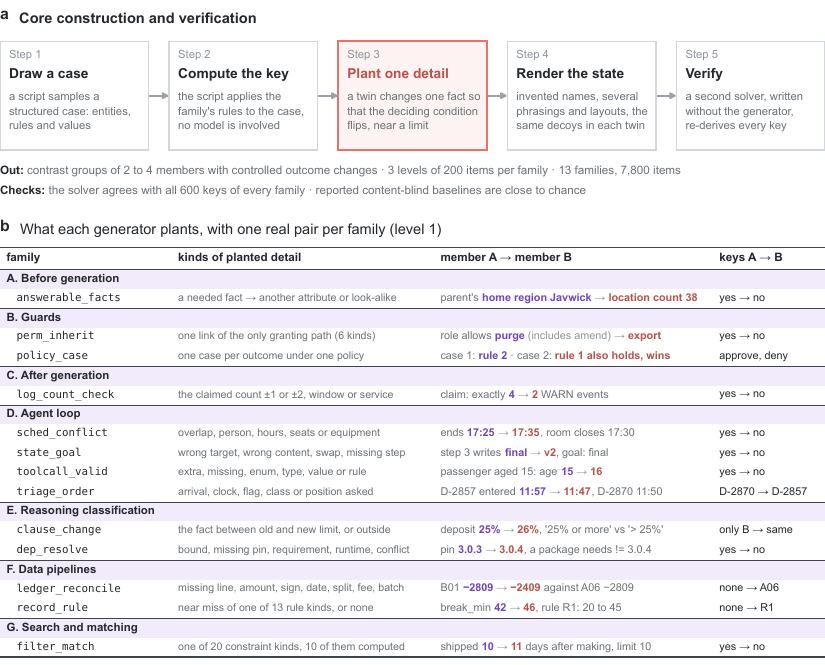}
\caption{\textbf{Core items combine controlled variations with independent answer verification.} (a) A generator samples a structured case, computes its answer, constructs contrast cases, and renders their states. An independently implemented solver verifies the computed answers. Most families vary a decisive fact while preserving the surrounding context. (b) One level-1 example from each family, showing member-A values in violet, member-B values in red, and their corresponding answers. In \texttt{policy\_case}, different cases share a policy; rule 1 also applies to the second example and takes precedence over rule 2.}\label{fig:3}
\end{figure}

For each core task family, a Python generator samples a structured case, computes its answer, and renders the state using invented entities and varied wording (Figure~\ref{fig:3}).
The released items are fixed for evaluation, while seven generators also support fresh samples drawn with configurable seeds.
Items are organized into contrast groups containing related cases with different correct answers.
In most families, a decisive fact is varied while preserving the surrounding context and distractors.
In \texttt{policy\_case}, cases with different outcomes instead share a common policy.
Appendix~\ref{sec:A} documents the contrast-group construction for each family.

For threshold-based decisions, positive and negative cases are constructed near the decision boundary.
For example, Box~\ref{box:1} illustrates how changing only the shipping fee moves the total price from one cent below to one cent above the specified limit.
Each family contains 600 items, evenly distributed across three difficulty levels (Table~\ref{tab:2}).
Higher levels increase case size or the complexity of rule application, including the number of entities, conditions, or required computations.
Because these factors vary together, the difficulty levels do not isolate context length from reasoning demands.

\begin{mdbox}{1}{box:1}
\textbf{Box~\ref{box:1}. A level-1 contrast pair from \texttt{filter\_\allowbreak{}match}, abridged.}
The two members differ only in the shipping fee, placing the total price one cent below or above the specified limit.

{\small\setlength{\tabcolsep}{6pt}\renewcommand{\arraystretch}{1.15}
\noindent\begin{tabular}{@{}>{\raggedright\arraybackslash}p{0.24\linewidth}>{\raggedright\arraybackslash}p{0.70\linewidth}@{}}
\toprule
\textbf{part} & \textbf{content} \\
\midrule
\mdbodyrow Request & The listing must satisfy every point below. Its warranty is not below 36 months. The price plus the shipping fee is less than 500.84 SEK. The price per unit is 415.77 SEK or less. \dots \\
\mdbodyrow Record & Lumrov Mini-404 by Melzim, price 476.80 SEK, 10 units for 4,157.70 SEK, warranty 37 months, 32 fields in all \\
\mdourrow Member A & shipping fee 24.03 SEK, price plus shipping 500.83 SEK, \textbf{answer yes} \\
\mdourrow Member B & shipping fee 24.05 SEK, price plus shipping 500.85 SEK, \textbf{answer no} \\
\midrule
\mdgrouprow \multicolumn{2}{@{}l@{}}{\textbf{Probability of yes for member A / member B}} \\
\mdbodyrow Perplexity Decider & 0.86 / 0.84 \\
\mdbodyrow SL-27B Instant & 0.90 / 0.83 \\
\mdbodyrow Clef 27B & 0.09 / 0.09 \\
\mdbodyrow GLM-5.3, lowest effort & 1.00 / 0.00 \\
\bottomrule
\end{tabular}\par}

None of the displayed one-pass models correctly answers both members.
GLM-5.3 answers both correctly at its lowest reasoning-effort setting, which still includes a short hidden reasoning trace.
\end{mdbox}

\begin{table}[!htbp]
\renewcommand{\thetable}{2}
\centering
\caption{\textbf{Core task families and reported content-blind baselines.} Contrast groups contain related cases with controlled outcome changes; \texttt{policy\_case} instead groups cases under a shared policy. The content-blind column reports baseline values from the task summaries, in percent; Appendix~\ref{sec:E} provides separate shortcut-audit results. State length is the median number of characters. Each family contains 200 items per difficulty level, and an independent solver agrees with all 600 answers. Higher levels increase case size or complexity (Appendix~\ref{sec:A}).}\label{tab:2}
{\small\setlength{\tabcolsep}{5pt}\renewcommand{\arraystretch}{1.1}
\noindent\begin{adjustbox}{max width=\linewidth}
\begin{tabular}{llrrrrr}
\toprule
\textbf{family} & \textbf{question} & \textbf{options} & {\begin{tabular}[b]{@{}r@{}}\textbf{contrast}\\\textbf{groups}\end{tabular}} & {\begin{tabular}[b]{@{}r@{}}\textbf{content-}\\\textbf{blind rule}\end{tabular}} & \textbf{chance} & {\begin{tabular}[b]{@{}r@{}}\textbf{state,}\\\textbf{characters}\end{tabular}} \\
\midrule
\mdgrouprow \multicolumn{7}{l}{\textbf{A. Before generation}} \\
\mdbodyrow answerable\_facts & can the facts answer the question & 2 & 300 & 50.7 & 50.0 & 2,470 \\
\midrule
\mdgrouprow \multicolumn{7}{l}{\textbf{B. Guards}} \\
\mdbodyrow perm\_inherit & may the user act on the folder & 2 & 300 & 50.3 & 50.0 & 3,926 \\
\mdbodyrow policy\_case & approve, deny or escalate & 3 & 201 & 33.3 & 33.3 & 1,703 \\
\midrule
\mdgrouprow \multicolumn{7}{l}{\textbf{C. After generation}} \\
\mdbodyrow log\_count\_check & is the claimed event count right & 2 & 300 & 50.5 & 50.0 & 2,291 \\
\midrule
\mdgrouprow \multicolumn{7}{l}{\textbf{D. Agent loop}} \\
\mdbodyrow sched\_conflict & can the booking be confirmed & 2 & 300 & 50.2 & 50.0 & 3,029 \\
\mdbodyrow state\_goal & does the trace reach the goal & 2 & 300 & 50.5 & 50.0 & 2,475 \\
\mdbodyrow toolcall\_valid & is the tool call valid & 2 & 300 & 50.7 & 50.0 & 4,975 \\
\mdbodyrow triage\_order & which case is at the asked place & 5 & 201 & 20.0 & 20.0 & 2,256 \\
\midrule
\mdgrouprow \multicolumn{7}{l}{\textbf{E. Reasoning classification}} \\
\mdbodyrow clause\_change & does the edit change the answer & 3 & 201 & 33.3 & 33.3 & 3,730 \\
\mdbodyrow dep\_resolve & do the pinned versions install & 2 & 300 & 50.3 & 50.0 & 3,348 \\
\midrule
\mdgrouprow \multicolumn{7}{l}{\textbf{F. Data pipelines}} \\
\mdbodyrow ledger\_reconcile & which entry fails, or none & 6 & 201 & 16.7 & 16.7 & 2,764 \\
\mdbodyrow record\_rule & which rule is broken, or none & 7 to 10 & 150 & 12.2 & 12.0 & 4,751 \\
\midrule
\mdgrouprow \multicolumn{7}{l}{\textbf{G. Search and matching}} \\
\mdbodyrow filter\_match & does the record pass every filter & 2 & 300 & 50.5 & 50.0 & 1,647 \\
\bottomrule
\end{tabular}
\end{adjustbox}\par}
\end{table}

\subsection{Reference track}\label{sec3.2.2}

The reference track comprises 35 public test sets spanning all seven decision layers (Table~\ref{tab:C1}).
Conversion scripts reconstruct evaluation items from official source files while preserving each task's original scoring metric.
We use fixed-seed sampling, targeting at most 1,000 items per task, except for two reranking tasks that target 100 queries with 50 candidates each.
Final item counts after filtering are reported in Table~\ref{tab:C2}.
Of the 35 tasks, 21 use human-annotated labels, six use programmatically computed labels, and eight draw on mixed label sources.
The two capability-boundary tasks use success-rate labels derived from sampled responses of specified public chat models, evaluating predicted success probabilities rather than correctness against a single gold option.%
\subsection{Quality, diversity and coverage}\label{sec:quality}

\textbf{Answer verification and review.}
We distinguish computational answer verification from textual quality review.
For each family, an author who had not read the generator independently implemented a solver that agrees with all 600 gold answers.
The three families unchanged since version 1.0 were also checked on a second draw (Table~\ref{tab:E1}).
Generator assertions check structural consistency and intended variations.
All 7,800 core items were independently screened by at least two human reviewers, with additional AI-assisted screening; flagged items were revised or removed.
Independent solvers verify agreement on computed answers, while textual review assesses whether each case is clearly specified (Appendix~\ref{sec:E}).

\textbf{Shortcut audits.}
We examine potential shortcuts using option-only, question-wording, state-length, line-count, and bag-of-words probes (Tables~\ref{tab:2} and \ref{tab:E1}).
Reported content-blind baselines remain close to chance; for example, the options-only probe achieves 21.3\% accuracy on \texttt{triage\_order}, compared with a 20.0\% chance baseline.
The seven families with configurable seeds balance answer classes and gold-option positions, and no core state is repeated.
However, partial-information heuristics can still succeed: checking only the schema achieves 91.7\% item accuracy on \texttt{toolcall\_valid} (Appendix~\ref{sec:E}).
These audits distinguish surface cues from shortcuts that exploit only part of a case, without ruling out all possible shortcuts.
Reference tasks retain their source artifacts, including the 0.730 content-blind baseline on \texttt{llmbar}.

\textbf{Revision and fresh-sample evaluation.}
Version 1.1, dated 8 October 2026, revises ten task families to address identified item defects (Table~\ref{tab:E2}).
The reported evaluation combines version-1.1 core items with the filtered version-1.3 reference track.
Across 20 configurations evaluated on core versions 1.0 and 1.1, rankings remain highly correlated (Spearman $\rho=0.986$).
Seven generators with configurable seeds also produced 4,200 fresh items after the initial evaluations.
For the 34 decision models included in this analysis, rankings across draws show a Spearman correlation of 0.994, and score-change confidence intervals include zero for 32 models (Table~\ref{tab:E3}).
These results assess sensitivity to fresh samples within the same generator designs, rather than generalization to independently designed cases.

\textbf{Training overlap.}
We removed 3,016 reference items (8.5\%) detected in the training files of our final 27B models, together with affected groups.
This filtering removes detected overlap for our models only.
For other models, we record whether public training documentation names a task source or related data family, marking missing or inconclusive information as unknown.
These source-level flags provide provenance context, not item-level proof of exposure or its absence (Appendix~\ref{sec:C}).

\textbf{Task coverage.}
Both tracks cover all seven decision roles, with their distributions shown in Figure~\ref{fig:2} and Table~\ref{tab:C1}.
Core items have two to ten options, with median state lengths ranging from 1,647 to 4,975 characters across families.
Recorded domain variation spans up to eight domains per family.
Decisive variations include twenty constraint types in \texttt{filter\_match} and thirteen integrity-rule types in \texttt{record\_rule} (Appendix~\ref{sec:A}).
\subsection{Provenance, licenses and use}\label{sec:license}

Core items contain no third-party text and are released under CC BY 4.0, while the accompanying code is licensed under Apache-2.0.
Reference items remain subject to their original source-specific terms.
The release combines evaluation data with reconstruction support, including scripts for 13 reference tasks. Appendix~\ref{sec:C} documents provenance, source-specific terms, and release routes.
Access conditions restrict the items to evaluation use.
Appendix~\ref{sec:F} provides a dataset datasheet following \citet{gebru2021datasheets}.
\subsection{Scoring}\label{sec:3.4}

\textbf{Requests and predictions.}
Each item is presented as a state, a typed question, and an ordered option set.
Decision models are evaluated through their documented interfaces; general-model prompting and probability readout are described in Appendix~\ref{sec:B}.
The option order is fixed in the evaluation files, and the predicted label is the option with the highest returned probability.
Failed or missing predictions count as incorrect rather than being excluded from the accuracy denominator.
Run validation checks item coverage and, for core families, the hash of the evaluated item file.
Model-specific input restrictions are documented in Table~\ref{tab:B1}.

\textbf{Group-level correctness.}
Let \(\mathcal G_f\) denote the set of contrast groups in family \(f\).
We define
\begin{equation}
\operatorname{PairAcc}_f=\frac{1}{|\mathcal G_f|}\sum_{g\in\mathcal G_f}\prod_{i\in g}\mathbf 1[\hat y_i=y_i].
\label{eq:pairacc}
\end{equation}
Pair accuracy requires every member of a contrast group to be answered correctly, including groups containing more than two items.
A change in prediction is insufficient if any group member remains incorrect.
For example, a model predicting yes for both members of Box~\ref{box:1} achieves 50\% item accuracy but zero pair accuracy.
Because all members must be correct, the chance baseline depends on group size and option count: independent uniform guessing succeeds with probability \(\prod_{i\in g}|O_i|^{-1}\).
We report the corresponding aggregate chance baseline alongside model scores.
Pair accuracy measures joint correctness but does not, by itself, identify why a group fails.

\textbf{Pair direction.}
For a binary contrast, define \(p^+=p(\texttt{yes}\mid s^+,q,O)\) and \(p^-=p(\texttt{yes}\mid s^-,q,O)\), where the respective gold labels are yes and no.
Pair direction measures the fraction of pairs satisfying \(p^+>p^-\).
Away from ties, correct predictions on both members require \(p^+>0.5>p^-\).
The probabilities 0.86 and 0.84 in Box~\ref{box:1} therefore yield correct direction but two yes predictions.
Pair-direction diagnostics cover 2,400 pairs from eight binary families (Table~\ref{tab:D6}), whereas the headline pair-accuracy metric covers all 13 families.
These diagnostics distinguish observed probability ordering from label correctness without identifying an internal reasoning mechanism.

\textbf{Confidence-based evaluation.}
Let \(c_i=\max_{o\in O_i}p(o\mid s_i,q_i,O_i)\), and order the \(n\) items in each family by decreasing confidence.
For a prefix of length \(k\), empirical risk and coverage are defined as
\begin{equation}
\widehat R(k)=\frac{1}{k}\sum_{j=1}^{k}\mathbf 1[\hat y_{(j)}\ne y_{(j)}],\qquad
\widehat C(\epsilon)=\max\left(\{0\}\cup\left\{\frac{k}{n}:1\le k\le n,\ \widehat R(k)\le\epsilon\right\}\right).
\label{eq:coverage}
\end{equation}
Eligible prefixes end at confidence-tie boundaries, so equal-confidence items enter together.
We report \(\widehat C(0.05)\) as \emph{automation at 5\%} in the tables.
This is retrospective coverage computed using evaluation labels.
Because coverage is computed separately for each family, the definition permits family-specific confidence cutoffs rather than a single threshold shared across tasks.
Selecting an operating threshold for future requests requires independent calibration.
We also report accuracy and coverage at confidence 0.90 and calibration error over 15 bins.

\begin{table}[!tbp]
\renewcommand{\thetable}{3}
\centering
\caption{\textbf{Metrics for correctness and probability quality.} The first three criteria assess predicted answers, while the remaining four evaluate probability quality. The predicted answer is the option assigned the highest probability. Abbreviated metric names are shown in parentheses.}\label{tab:3}
{\footnotesize\setlength{\tabcolsep}{5pt}\renewcommand{\arraystretch}{1.1}
\noindent\begin{adjustbox}{max width=\linewidth}
\begin{tabular}{>{\raggedright\arraybackslash}p{\mdw{0.24}}>{\raggedright\arraybackslash}p{\mdw{0.42}}>{\raggedright\arraybackslash}p{\mdw{0.34}}}
\toprule
\textbf{criterion} & \textbf{what it measures} & \textbf{what it catches} \\
\midrule
\mdgrouprow \multicolumn{3}{l}{\textbf{The answers}} \\
\mdbodyrow accuracy & items whose top option is correct & incorrect predictions \\
\mdbodyrow pair accuracy & contrast groups with every member correct & failure to distinguish related outcomes: predicting yes for both members of Box~\ref{box:1} gives item accuracy 50 and pair accuracy 0 \\
\mdbodyrow level 3 & pair accuracy at the highest difficulty level & performance decline as case size or complexity increases \\
\midrule
\mdgrouprow \multicolumn{3}{l}{\textbf{The probabilities}} \\
\mdbodyrow pair direction & binary contrast pairs whose positive member receives a higher probability of yes & correct relative ordering without necessarily correct predictions on both members \\
\mdbodyrow automation at 5\% (auto at 5\%) & maximum empirical coverage of a confidence-ranked prefix with error at most 5\% & limited separation of correct and incorrect predictions by confidence \\
\mdbodyrow accuracy at 0.90 (acc. at 0.90) & accuracy among items with top probability at least 0.90, and their share & high-confidence errors \\
\mdbodyrow calibration error & mean gap between predicted confidence and observed accuracy over 15 bins \citep{guo2017calibration} & overconfidence or underconfidence \\
\bottomrule
\end{tabular}
\end{adjustbox}\par}
\end{table}

\textbf{Aggregation and uncertainty.}
Core headline scores are macro-averaged over 13 families; Table~\ref{tab:D1} identifies confidence metrics computed on pooled items.
For reference task \(t\), a score \(v_t\) with chance baseline \(b_t\) is transformed to \(100(v_t-b_t)/(1-b_t)\).
Scores are floored at zero before within-layer averaging, and the overall reference score averages the seven layers; Table~\ref{tab:D4} retains negative task scores for inspection.
We report 95\% percentile bootstrap intervals, resampling groups where available and individual items otherwise, with matched resamples for paired comparisons (Appendix~\ref{sec:B.4}).
Uniform guessing yields 16.1\% core pair accuracy.
%
\FloatBarrier
\section{Experiments}\label{sec:4}

Our experiments investigate three aspects of decision-model performance: what controlled contrasts reveal beyond aggregate scores, how confidence relates to prediction correctness, and how performance changes across reasoning settings.
We evaluate 40 decision models, including two StartLux-Decision-27B variants \citep{startlux2026report}, 37 open models from 20 other groups, and jev-1.13.0 (the \emph{hosted reference}).
Instant operates in a single pass, while AutoThink generates a reasoning trace when its highest one-pass probability falls below 0.9.
The six Decision 3.0 models are included only in the main comparison.
% Model-release provenance: the six Decision 3.0 models were released on 10 October 2026 and evaluated after the other models.
We additionally evaluate 11 open general models using one-pass option-key probabilities on both tracks and three API chat models using stated probabilities on the core track (Table~\ref{tab:Dgen}).
Appendix~\ref{sec:B} details model revisions, inference interfaces, and input restrictions.

\subsection{Main results}\label{sec:4.1}

\begin{table}[!htbp]
\renewcommand{\thetable}{4}
\centering
\caption{\textbf{Main results for 40 decision models.} Core metrics are reported as percentages, macro-averaged over 13 task families. Models are sorted by pair accuracy, with 95\% bootstrap intervals for that metric. Level 3 denotes the highest difficulty level; auto at 5\% measures retrospective confidence-ranked coverage at a 5\% empirical error rate. The reference score aggregates 35 public tasks (0 = chance). Uniform guessing yields 39.6\% item accuracy and 16.1\% pair accuracy. $^\dagger$ Input restrictions cause missing or truncated predictions; missing predictions count as incorrect (Table~\ref{tab:B1}). $^\ddagger$ Decision 3.0 models appear only in this main comparison. Table~\ref{tab:Dgen} reports general models.}\label{tab:4}
{\small\setlength{\tabcolsep}{6pt}\renewcommand{\arraystretch}{1.1}
\noindent\begin{adjustbox}{max width=\linewidth}
\begin{tabular}{lllrrrrr}
\toprule
\textbf{decision model} & \textbf{organization} & \textbf{size} & \textbf{accuracy} & \textbf{pair accuracy} & {\begin{tabular}[b]{@{}r@{}}\textbf{level}\\\textbf{3}\end{tabular}} & {\begin{tabular}[b]{@{}r@{}}\textbf{auto}\\\textbf{at 5\%}\end{tabular}} & \textbf{reference} \\
\midrule
\mdourrow \textbf{SL-27B AutoThink} & StartLux (ours) & 27B & 94.0 & 86.9 {[}85.7, 88.0{]} & 82.2 & 93.8 & 58.9 \\
\mdourrow \textbf{SL-27B Instant} & StartLux (ours) & 27B & 86.4 & 75.4 {[}74.1, 76.7{]} & 65.3 & 79.6 & 59.4 \\
\mdbodyrow Decision 3.0 d3$^\ddagger$ & vLLM Semantic Router & 27B & 83.9 & 71.3 {[}69.9, 72.8{]} & 59.0 & 75.1 & 56.7 \\
\mdbodyrow Perplexity Decider v1.1 & Perplexity & 27B & 83.4 & 70.1 {[}68.7, 71.5{]} & 57.4 & 72.3 & 57.0 \\
\mdbodyrow Decision 2.0 Vega & vLLM Semantic Router & 27B & 73.7 & 51.4 {[}49.7, 53.1{]} & 33.3 & 48.4 & 55.5 \\
\mdbodyrow Kev 27B & Jared Palmer & 27B & 70.6 & 44.2 {[}42.6, 46.0{]} & 26.5 & 35.2 & 54.0 \\
\mdbodyrow JEV-27B & AutoTrust AI Lab & 27B & 70.3 & 44.2 {[}42.6, 45.9{]} & 27.0 & 34.8 & 54.4 \\
\mdbodyrow Jebadiah 27B & Frontier Infra & 27B & 69.8 & 43.3 {[}41.6, 45.0{]} & 25.4 & 32.3 & 55.3 \\
\mdbodyrow Decision 3.0 d3-flash$^\ddagger$ & vLLM Semantic Router & 9B & 67.6 & 43.3 {[}41.8, 44.9{]} & 27.6 & 36.9 & 52.6 \\
\mdbodyrow Decision 3.0 d3-mini$^\ddagger$ & vLLM Semantic Router & 4B & 64.3 & 38.0 {[}36.5, 39.6{]} & 21.9 & 23.3 & 51.3 \\
\mdbodyrow Jev 1.13 (hosted) & TypeSafe & -- & 63.9 & 31.0 {[}29.4, 32.7{]} & 15.9 & 19.3 & 55.0 \\
\mdbodyrow Blink v0.3 & Pixilab.ai & 26B-A4B & 60.4 & 27.3 {[}25.6, 28.9{]} & 12.3 & 11.3 & 52.6 \\
\mdbodyrow Xor 1.0 & Juspay & 26B-A4B & 59.8 & 26.2 {[}24.7, 27.7{]} & 12.0 & 14.5 & 53.4 \\
\mdbodyrow GEV-26B-Decide & AutoTrust AI Lab & 26B-A4B & 59.1 & 25.2 {[}23.7, 26.7{]} & 11.1 & 9.0 & 50.9 \\
\mdbodyrow Clef & Cloudflare & 27B & 60.0 & 24.9 {[}23.3, 26.6{]} & 12.8 & 14.5 & 52.4 \\
\mdbodyrow Hopper 12B & Hopit & 12B & 55.2 & 20.5 {[}19.0, 21.8{]} & 9.0 & 5.9 & 51.1 \\
\mdbodyrow H2O-Lightning-4B & H2O.ai & 4B & 55.4 & 19.7 {[}18.4, 21.0{]} & 10.3 & 5.9 & 47.0 \\
\mdbodyrow Decision 2.0 Lux & vLLM Semantic Router & 9B & 54.1 & 19.6 {[}18.3, 20.8{]} & 11.1 & 5.1 & 50.3 \\
\mdbodyrow Nimble v3 & Bespoke Labs & 9B & 54.7 & 18.3 {[}17.0, 19.6{]} & 8.9 & 5.9 & 50.1 \\
\mdbodyrow Decision 2.0 Nox & vLLM Semantic Router & 4B & 52.0 & 16.4 {[}15.1, 17.6{]} & 7.8 & 1.8 & 47.6 \\
\mdbodyrow JEV-9B & AutoTrust AI Lab & 9B & 52.4 & 16.1 {[}14.8, 17.3{]} & 7.4 & 2.8 & 48.7 \\
\mdbodyrow FlyMy Decision 4B v1.2$^\dagger$ & FlyMy.AI & 4B & 49.7 & 13.3 {[}12.3, 14.4{]} & 7.3 & 3.5 & 43.9 \\
\mdbodyrow Eval Engine Decision-4B$^\dagger$ & Eval Engine (Chromia) & 4B & 43.9 & 12.9 {[}11.8, 14.1{]} & 3.9 & 2.3 & 40.4 \\
\mdbodyrow Clef flash & Cloudflare & 9.5B & 50.7 & 12.5 {[}11.5, 13.6{]} & 6.2 & 3.0 & 48.2 \\
\mdbodyrow CoCo-Decision-4B-Ko$^\dagger$ & Corners & 4B & 49.1 & 11.8 {[}10.7, 12.8{]} & 5.5 & 2.0 & 47.4 \\
\mdbodyrow NeoHorse-Jev-4B$^\dagger$ & TokenRhythm & 4B & 45.2 & 11.8 {[}10.8, 12.9{]} & 5.2 & 1.4 & 42.1 \\
\mdbodyrow Decision 3.0 d3-nano$^\ddagger$ & vLLM Semantic Router & 2B & 48.1 & 11.7 {[}10.6, 12.8{]} & 5.3 & 1.2 & 44.8 \\
\mdbodyrow Hopper (G) 1.2 & Hopit & 4B & 49.5 & 10.5 {[}9.5, 11.6{]} & 3.8 & 3.1 & 47.1 \\
\mdbodyrow Intern-Decision-4B$^\dagger$ & InternLM & 4B & 48.0 & 9.8 {[}8.9, 10.7{]} & 5.0 & 3.2 & 43.9 \\
\mdbodyrow Decision 3.0 d3-lite$^\ddagger$ & vLLM Semantic Router & 0.8B & 44.3 & 7.4 {[}6.6, 8.3{]} & 4.0 & 0.6 & 37.7 \\
\mdbodyrow Intern-Decision-2B$^\dagger$ & InternLM & 2B & 43.7 & 7.1 {[}6.4, 7.9{]} & 6.2 & 0.4 & 32.8 \\
\mdbodyrow d1-3B & Liquid AI & 3B & 43.7 & 6.8 {[}6.0, 7.6{]} & 4.8 & 0.2 & 45.3 \\
\mdbodyrow Decision 2.0 Sol & vLLM Semantic Router & 2B & 43.3 & 6.2 {[}5.6, 7.0{]} & 4.8 & 0.4 & 33.7 \\
\mdbodyrow this-that 1.2$^\dagger$ & FLock.io & 2B & 43.1 & 6.0 {[}5.3, 6.9{]} & 2.8 & 0.2 & 36.0 \\
\mdbodyrow Strands decider 2B v21 & Strands Agents (AWS) & 2B & 42.3 & 5.5 {[}4.8, 6.3{]} & 4.8 & 0.1 & 34.4 \\
\mdbodyrow Decision 2.0 Eos & vLLM Semantic Router & 0.8B & 41.8 & 4.4 {[}3.7, 5.1{]} & 2.7 & 0.2 & 24.8 \\
\mdbodyrow Bosun 3.1 1.7B & Hanno Labs & 1.7B & 40.9 & 4.1 {[}3.4, 4.7{]} & 4.2 & 0.1 & 27.7 \\
\mdbodyrow d1-omni-600M & Liquid AI & 0.6B & 39.8 & 3.7 {[}3.2, 4.3{]} & 3.2 & 0.1 & 16.2 \\
\mdbodyrow Decision 2.0 Kai$^\dagger$ & vLLM Semantic Router & 0.6B & 40.6 & 3.6 {[}3.1, 4.2{]} & 2.6 & 0.1 & 25.2 \\
\mdbodyrow Decision 3.0 d3-edge$^\ddagger$ & vLLM Semantic Router & 0.6B & 41.1 & 3.5 {[}2.9, 4.1{]} & 2.4 & 0.2 & 28.6 \\
\bottomrule
\end{tabular}
\end{adjustbox}\par}
\end{table}

\begin{figure}[!tbp]
\renewcommand{\thefigure}{4}
\centering
\adjincludegraphics[max width=\linewidth,max height=0.5\textheight]{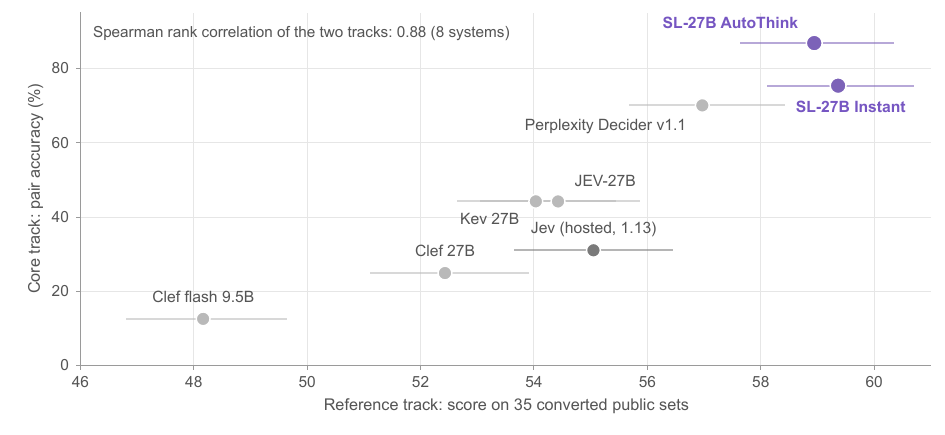}
\caption{\textbf{Core and reference rankings broadly agree.} Core pair accuracy versus reference aggregate score for eight selected models, with 95\% intervals. Spearman rank correlation is 0.88. Score ranges are not directly comparable because the tracks differ in task content and scoring.}\label{fig:4}
\end{figure}

\textbf{Similar reference scores, different core performance.}
Perplexity Decider and JEV-27B achieve reference scores of 57.0 and 54.4, respectively, yet differ substantially in core pair accuracy (70.1\% versus 44.2\%; Table~\ref{tab:4}).
Across the eight models in Figure~\ref{fig:4}, rankings broadly agree (Spearman $\rho=0.88$), despite ranges of 11 reference-score points and 74 percentage points in core pair accuracy.
Because the tracks differ in both task content and scoring, these differences cannot be attributed to group-level evaluation alone.

\textbf{Complete contrast groups remain challenging for many models.}
Among one-pass decision models, Instant achieves 75.4\% pair accuracy, followed by Decision 3.0 d3 at 71.3\% and Perplexity Decider at 70.1\%.
Of the 40 evaluated decision models, 19 fall below the 16.1\% uniform-guess baseline in overall pair accuracy, while 30 fall below 16.1\% at difficulty level 3.
These results include failures caused by input restrictions and therefore reflect both input compatibility and decision performance.
Performance also varies substantially across task families: Perplexity Decider achieves 97.5\% pair accuracy on \texttt{policy\_case}, compared with 33.8\% on \texttt{triage\_order} (Table~\ref{tab:D2}).

\textbf{Reference tasks provide complementary comparisons.}
On the reference track, Instant achieves a score of 59.4, exceeding Perplexity Decider by 2.4 points [1.6, 3.2] in the paired comparison.
Its point estimate is higher in six of the seven decision layers, although confidence intervals for some layer-wise differences include zero (Table~\ref{tab:D3}).
Thus, the aggregate advantage does not establish superior performance across every decision role.

\subsection{Reliability beyond accuracy}\label{sec:4.2}

\textbf{Similar accuracy, different confidence reliability.}
Kev 27B and JEV-27B achieve nearly identical core item accuracy (70.6\% and 70.3\%, respectively), yet differ in the reliability of their high-confidence predictions.
Among predictions with confidence at least 0.90, Kev 27B achieves 93.8\% accuracy on 34.1\% of items, compared with 88.9\% accuracy on 47.4\% of items for JEV-27B (Table~\ref{tab:D1}).
JEV-27B produces high-confidence predictions more frequently, but a larger proportion of these predictions are incorrect.
This distinction is not captured by overall item accuracy.

\begin{figure}[!htbp]
\renewcommand{\thefigure}{5}
\centering
\adjincludegraphics[max width=\linewidth]{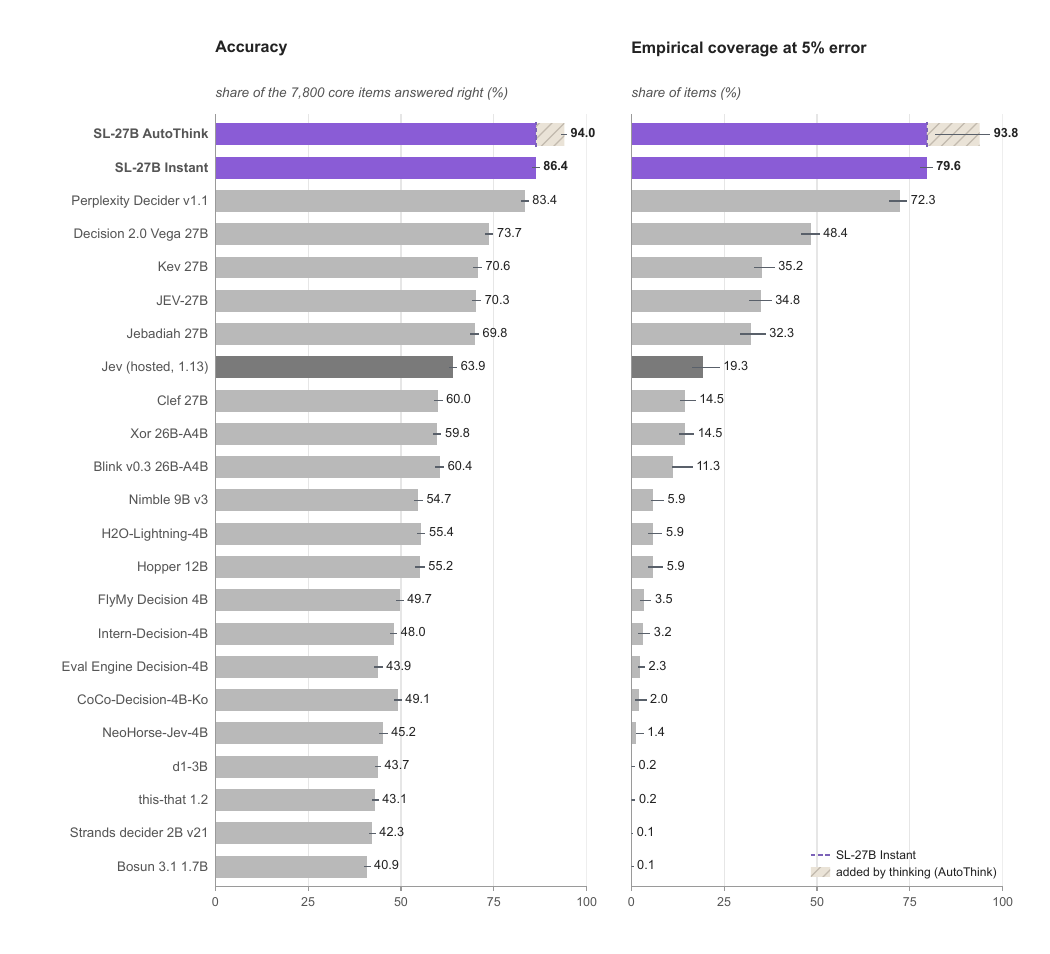}
\caption{\textbf{Confidence-based coverage differs from item accuracy.} Left: core item accuracy. Right: retrospective coverage of confidence-ranked predictions at a 5\% empirical error rate, macro-averaged across families. Models with approximately 70\% item accuracy achieve coverage of only about one-third of items at this error target. The plotted subset follows Figure~\ref{fig:D2}.}\label{fig:5}
\end{figure}

\textbf{Selective coverage provides a complementary view of confidence reliability.}
At a common empirical error target of 5\%, Kev 27B and JEV-27B achieve similar family-averaged coverage of 35.2\% and 34.8\%, respectively.
Perplexity Decider and Instant achieve substantially higher coverage of 72.3\% and 79.6\%, respectively, with non-overlapping confidence intervals (Figure~\ref{fig:5}).
Thus, similar selective coverage can coexist with different accuracy and prediction frequency at a fixed confidence threshold, highlighting the complementary information provided by these metrics.

\FloatBarrier%
\subsection{Performance with reasoning}\label{sec:4.3}

Higher reasoning settings are associated with improved performance on sampled core cases.
On a 1,296-item subset containing complete contrast groups with balanced difficulty levels, GLM-5.3 achieves 93.5\% [91.7, 95.3] pair accuracy at the higher setting, compared with 81.5\% [78.4, 84.6] at its lowest setting (Table~\ref{tab:D5}).
Average reasoning length increases from 121 to 265 tokens per item, although even the lowest setting involves hidden reasoning.
On the same subset, AutoThink achieves 85.9\% pair accuracy with an average of 297 reasoning tokens per item, compared with 75.6\% for Instant and 70.4\% for Perplexity Decider.
The higher-effort chat runs also use a larger output budget and the provider's default sampling, whereas the lower-effort runs use temperature 0 (Appendix~\ref{sec:B.3}).
These comparisons therefore reflect differences in sampling and output budget as well as reasoning effort.

Across the full evaluation tracks, AutoThink achieves 86.9\% core pair accuracy, compared with 75.4\% for Instant, while their reference scores are 58.9 [57.7, 60.3] and 59.4, respectively.
AutoThink generates a reasoning trace on approximately two-thirds of reference items.
These results highlight that improvements on controlled contrasts do not necessarily translate into higher aggregate scores on established public tasks.

Aggregate stability can also conceal changes in individual predictions.
A second AutoThink run achieves 87.2\% core pair accuracy, close to the original 86.9\%, while changing 5.6\% of individual answers.
Thus, similar aggregate scores do not necessarily imply stable item-level decisions.%
\FloatBarrier
\section{Diagnostic analysis}\label{sec:5}

\begin{figure}[!tbp]
\renewcommand{\thefigure}{6}
\centering
\adjincludegraphics[max width=\linewidth,max height=0.5\textheight]{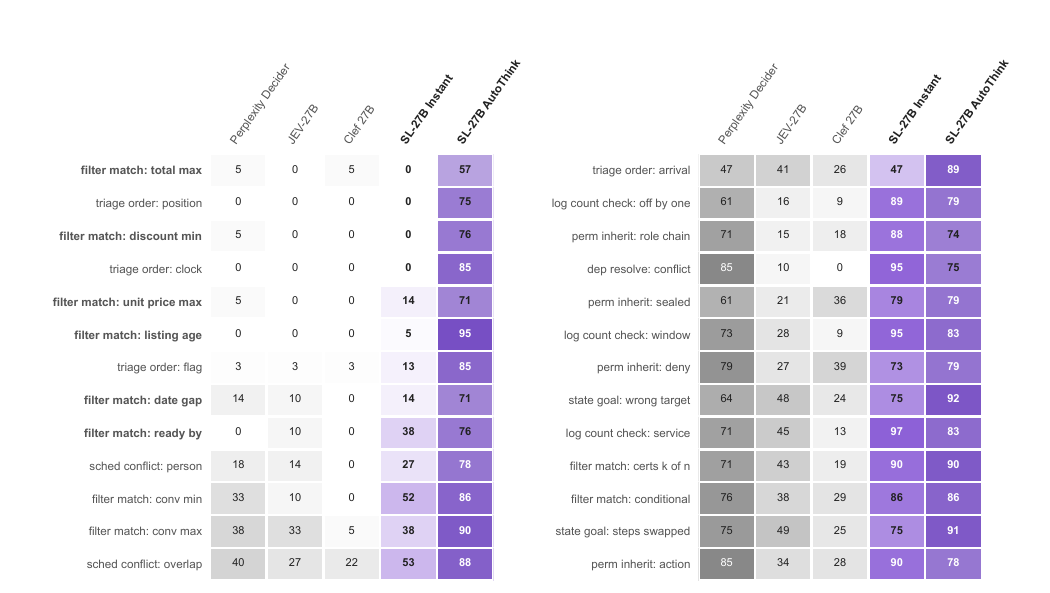}
\caption{\textbf{Selected arithmetic and temporal contrasts expose large performance gaps.} Pair accuracy for the 26 variation types with the lowest mean across the five displayed models, restricted to types with at least 20 groups. The final column shows SL-27B AutoThink with adaptive reasoning. Bold labels mark the six \texttt{filter\_\allowbreak{}match} variation types requiring arithmetic on prices or dates.}\label{fig:7}
\end{figure}

\begin{figure}[!tbp]
\renewcommand{\thefigure}{7}
\centering
\adjincludegraphics[max width=\linewidth,max height=0.5\textheight]{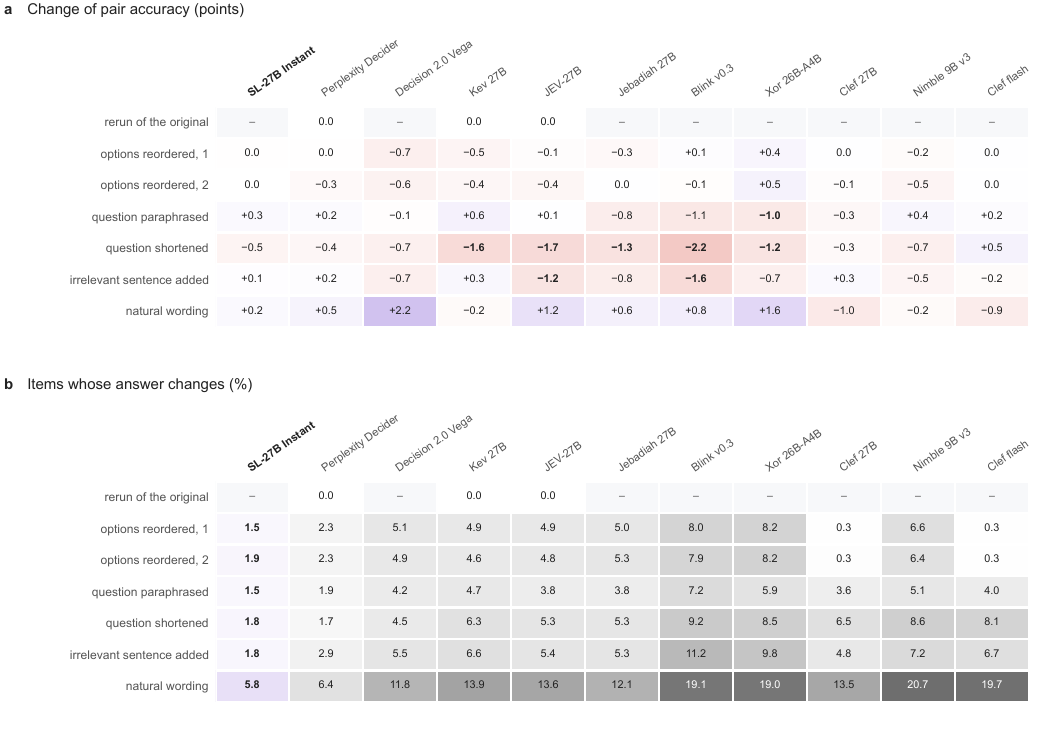}
\caption{\textbf{Small score changes can conceal changes in individual predictions.} Version-1.0 input variants for the displayed models. (a) Pair-accuracy changes, ranging from $-$2.2 to +2.2 percentage points. (b) Percentage of predictions that change, reaching 20.7\% for natural wording. Variant names follow the figure labels. Dashes denote unavailable entries; zero-valued reruns refer only to models with reported rerun results. Tables~\ref{tab:D8} and \ref{tab:D9} describe the four-model input-sensitivity analyses.}\label{fig:8}
\end{figure}

We use difficulty labels and annotations of decisive variations across 12 task families to examine where decision models fail.
Our analysis distinguishes probability ordering from label correctness, examines errors by case type and difficulty, and tests whether aggregate scores reflect changes in individual predictions.

\textbf{Correct probability ordering can coexist with incorrect predictions.}
Across 2,400 contrast pairs from eight binary task families, Perplexity Decider assigns a higher probability of yes to the positive member in 92.3\% of pairs, yet correctly classifies both members in only 74.1\%.
Clef 27B achieves 78.6\% pair direction but only 25.0\% pair accuracy (Table~\ref{tab:D6}).
As illustrated in Box~\ref{box:1}, correctly ordered probabilities can still fall on the same side of the 0.5 decision boundary.
Selecting a single global threshold on the evaluation pairs increases pair accuracy to 74.2\% and 30.6\% for the two models, respectively.
Even this optimistic choice closes only part of the gap and does not establish held-out calibration performance or the effectiveness of more flexible calibration methods.

\textbf{Error asymmetry varies across models.}
Across the 12 annotated families, Kev 27B achieves 56.5\% accuracy on cases marked ``no planted problem'' and 77.5\% on those with a planted problem, while Clef 27B achieves 32.7\% and 76.7\%, respectively (Table~\ref{tab:D7}).
In contrast, Perplexity Decider achieves 80.9\% accuracy in both categories.
These differences motivate evaluating both valid and invalid cases under a specified distribution before interpreting errors as false rejections or false acceptances in guard applications.

\textbf{Arithmetic and temporal contrasts expose specific weaknesses.}
Figure~\ref{fig:7} highlights substantial performance differences on selected arithmetic and temporal variations.
The four displayed one-pass models achieve at most 5\% pair accuracy on total-price and minimum-discount contrasts, compared with 57\% and 76\% for AutoThink.
On current-time contrasts in \texttt{triage\_order}, all four one-pass models score 0\%, whereas AutoThink reaches 85\%.
The figure selects the 26 variation types with the lowest mean performance across the five displayed models.
These results characterize selected difficult cases rather than average effects across all tasks or models.

\textbf{Performance declines as case size and complexity increase.}
From level 1 to level 3, Perplexity Decider's pair accuracy falls from 87.7\% to 57.4\%, while Instant's falls from 90.7\% to 65.3\%.
AutoThink shows a smaller decline, from 89.2\% to 82.2\% (Table~\ref{tab:D6}; Figure~\ref{fig:D1}).
Because difficulty levels increase both case size and complexity, these differences do not isolate either factor.

\textbf{Models differ in confidence assigned to errors.}
Among incorrect predictions, Perplexity Decider assigns confidence of at least 0.90 to 33.2\%, compared with 14.6\% for Instant (Table~\ref{tab:D7}).
This measures the proportion of errors made with high confidence, whereas Section~\ref{sec:4.2} examines correctness among high-confidence predictions.

\textbf{Aggregate stability can conceal changes in individual predictions.}
In the version-1.0 option-order study, Kev 27B changes 4.9\% of predictions under the first option-reordering variant.
Correct-to-incorrect changes affect 2.0\% of items, while incorrect-to-correct changes affect 1.5\%, resulting in a net accuracy decline of only 0.5 percentage points (Table~\ref{tab:D9}).
Thus, small aggregate score changes can mask more frequent changes in individual predictions.
Across the broader input-presentation study, pair-accuracy changes range from $-$2.2 to +2.2 points, while natural wording changes up to 20.7\% of predictions (Figure~\ref{fig:8}).
These analyses use version-1.0 items and do not establish robustness on the revised version 1.1.

Together, these analyses distinguish probability-ordering errors from incorrect decisions, identify weaknesses across case types and difficulty levels, and reveal changes in individual predictions.
Combined with confidence-based coverage, they provide a more detailed account of decision-model behavior than aggregate rankings alone.%
\FloatBarrier
\section{Limitations}\label{sec:6}

\textbf{Scope and realism.}
DecisionPivot evaluates written cases across seven decision roles rather than end-to-end system operation.
Generated cases prioritize controlled, verifiable answers, while real-world deployment, multilingual performance, and inference efficiency remain outside the current scope.
Difficulty levels vary case size and complexity together.

\textbf{Evaluation protocols and uncertainty.}
Models are evaluated through their documented interfaces, which differ in probability readout, input limits, and reasoning settings.
Most evaluations use a single run, and input-sensitivity analyses are based on version-1.0 items.
Empirical selective coverage is retrospective and requires independent calibration for deployment.
Reference tasks retain their original metrics, with potential uncertainty in source labels.

\textbf{Data exposure and benchmark design.}
Training-data overlap cannot be fully assessed for models without detailed training records; our filtering addresses detected overlap in our own models.
Fresh-sample evaluations assess stability within the existing generator designs rather than across independently constructed cases.
As DecisionPivot and StartLux-Decision are developed by the same team, independent evaluation and additional case designs would provide valuable complementary evidence.%
\FloatBarrier
\section{Conclusion}\label{sec:7}

We introduce DecisionPivot, a diagnostic benchmark that combines controlled contrast groups with independently verified answers and complementary public evaluation tasks.
By jointly assessing item correctness, contrast-group correctness, probability ordering, and confidence reliability, DecisionPivot provides a more comprehensive view of decision-model performance.
Our evaluation reveals that models with similar reference scores can differ substantially on controlled contrasts, while stable aggregate scores can mask changes in individual predictions.
Comparisons across reasoning settings further reveal improvements on sampled core cases, with different performance patterns across evaluation tracks.
Together, these findings highlight the value of fine-grained diagnostics for identifying decision failures and guiding the evaluation of reliable decision models.%
\FloatBarrier
\mdBibliography
\clearpage
\appendix
\section*{Appendix contents}
\markboth{Appendix contents}{}
\begingroup
\setlength{\parskip}{0pt}
\newcommand{\mdAppendixContentsEntry}[1]{%
  \par\noindent
  \hyperref[#1]{\makebox[2em][l]{\ref*{#1}.}\nameref*{#1}}%
  \nobreak\hspace{0.5em}\dotfill\nobreak\hspace{0.5em}%
  \hyperref[#1]{\pageref*{#1}}\par
  \vspace{0.6em}%
}
\mdAppendixContentsEntry{sec:A}
\mdAppendixContentsEntry{sec:B}
\mdAppendixContentsEntry{sec:C}
\mdAppendixContentsEntry{sec:D}
\mdAppendixContentsEntry{sec:E}
\mdAppendixContentsEntry{sec:F}
\endgroup
\clearpage
\FloatBarrier
\section{Task cards of the 13 core families}\label{sec:A}
\begingroup
\setlength{\parskip}{0.25em plus 0.1em minus 0.1em}

Each family contains 600 items, with 200 per difficulty level and a shared instruction.
The generator computes each answer from a structured case and renders the state using invented entities, varied wording, and explicit task conventions.
Answer classes and gold-option positions are balanced in the seven families with configurable seeds.
Distractors appear in each class, and members of a contrast group have the same number of lines.
Generator assertions check intended violations and structural consistency, while textual ambiguity is assessed separately through review.
In six families (\texttt{answerable\_\allowbreak{}facts}, \texttt{policy\_case}, \texttt{log\_\allowbreak{}count\_\allowbreak{}check}, \texttt{state\_goal}, \texttt{toolcall\_\allowbreak{}valid}, \texttt{filter\_\allowbreak{}match}), difficulty levels determine case size.
Each generator also produces a task README with one complete example per level, and Table~\ref{tab:E1} summarizes the shared checks.

\subsection{\texorpdfstring{\texttt{answerable\_\allowbreak{}facts} (layer A)}{answerable facts (layer A)}}\label{sec:A.1}

\textbf{Question.} ``Can the question be answered from the facts listed?'' (\texttt{yes}, \texttt{no}).
\textbf{State.} A table of facts about invented organizations, interpretation rules, and a question requiring two to four rows.
\textbf{Levels.} Two required facts among 20 rows, three among 40, and four among 80.
\textbf{Groups.} The negative member replaces one required row with another attribute or a look-alike organization (150 pairs each).

\subsection{\texorpdfstring{\texttt{perm\_\allowbreak{}inherit} (layer B)}{perm inherit (layer B)}}\label{sec:A.2}

\textbf{Question.} ``Is the request allowed?'' (\texttt{yes}, \texttt{no}).
\textbf{State.} The written rules of an invented access system, including users, groups, roles, a folder tree, and an access request.
\textbf{Levels.} From a tree of depth 2 to a tree of depth 4 with nested groups, sealed folders, and deny entries.
\textbf{Groups.} The negative member breaks exactly one link in the only path granting access.
\textbf{Notes.} Combining fourteen hand-crafted structural counts yields 0.561 accuracy.

\subsection{\texorpdfstring{\texttt{policy\_case} (layer B)}{policy case (layer B)}}\label{sec:A.3}

\textbf{Question.} ``What does the policy require for this case?'' (\texttt{approve}, \texttt{deny}, \texttt{escalate}).
\textbf{State.} A numbered policy containing 4, 8, or 12 rules with exceptions, a precedence statement, a default outcome, and a case record.
\textbf{Levels.} The number of rules.
\textbf{Groups.} 201 policies, most with one case per outcome, balanced by outcome, difficulty level, and required interpretation (plain, exception, or precedence statement; 200 items each).

\subsection{\texorpdfstring{\texttt{log\_\allowbreak{}count\_\allowbreak{}check} (layer C)}{log count check (layer C)}}\label{sec:A.4}

\textbf{Question.} ``Is the claim correct for this log?'' (\texttt{yes}, \texttt{no}).
\textbf{State.} A one-day event log containing 30, 60, or 90 lines, together with a claim about the exact number of events matching a specified service and level within a time window.
\textbf{Levels.} The number of log lines.
\textbf{Groups.} The negative member changes one field of a true claim: the count by one or two, the time window, or the service (75 pairs each).

\subsection{\texorpdfstring{\texttt{sched\_\allowbreak{}conflict} (layer D)}{sched conflict (layer D)}}\label{sec:A.5}

\textbf{Question.} ``Can the requested booking be confirmed as it stands?'' (\texttt{yes}, \texttt{no}).
\textbf{State.} Rooms with seating capacities, equipment, and opening hours; booking conventions; confirmed bookings; and a booking request.
\textbf{Levels.} From 2 rooms and 6 to 8 bookings to 5 to 6 rooms and 20 to 28 bookings across 2 to 3 days.
\textbf{Groups.} The positive member lies exactly at a limit that the negative member crosses, most often through a booking overlap (112 pairs) or a double-booked person (77).
\textbf{Notes.} The heuristic ``the room is exactly full, so yes'' achieves 0.665 accuracy at level 1.

\subsection{\texorpdfstring{\texttt{state\_goal} (layer D)}{state goal (layer D)}}\label{sec:A.6}

\textbf{Question.} ``After these operations, are all goal conditions satisfied?'' (\texttt{yes}, \texttt{no}).
\textbf{State.} A small file store from one of five domains, a description of operation semantics, a goal with 3, 5, or 7 conditions, and a numbered operation trace.
\textbf{Levels.} Traces of 5, 15, or 25 steps.
\textbf{Groups.} A single changed step causes exactly one goal condition to fail, through a wrong target, wrong content, swapped steps, or a replaced step (75 pairs each).

\subsection{\texorpdfstring{\texttt{toolcall\_\allowbreak{}valid} (layer D)}{toolcall valid (layer D)}}\label{sec:A.7}

\textbf{Question.} ``Is the proposed call correct for the request under this schema?'' (\texttt{yes}, \texttt{no}).
\textbf{State.} A JSON tool schema with 6, 9, or 12 top-level properties and rules in its description, a user request specifying values directly or indirectly, and the proposed call.
\textbf{Levels.} The number of properties.
\textbf{Groups.} One planted deviation per negative member, with 50 items per deviation type.

\subsection{\texorpdfstring{\texttt{triage\_\allowbreak{}order} (layer D)}{triage order (layer D)}}\label{sec:A.8}

\textbf{Question.} ``Which waiting case do the rules put at the position asked for in the state?''
Options are five tickets from the queue.
\textbf{State.} The current time, numbered ordering rules, the queue, and the requested position.
\textbf{Levels.} Level 1 contains 6 cases and 2 to 3 rules, while level 3 contains 12 to 14 cases, 5 to 6 rules, and two class adjustments.
\textbf{Groups.} Members differ in one element, most often an arrival time (217 items) or the current time (119).
\textbf{Notes.} The heuristic ``lowest listed class, then earliest arrival'' identifies the correct answer for 12.5\% and 7.5\% of items at levels 2 and 3, respectively.

\subsection{\texorpdfstring{\texttt{clause\_\allowbreak{}change} (layer E)}{clause change (layer E)}}\label{sec:A.9}

\textbf{Question.} ``Does the change between the two versions alter the answer to the point at issue?'' (\texttt{same}, \texttt{yes\_only\_A}, \texttt{yes\_only\_B}; 200 items each).
\textbf{State.} Interpretation rules, two versions of an agreement, and a case with facts.
\textbf{Levels.} From 8 clauses per version to 12 to 14 clauses.
\textbf{Groups.} Members place a fact between the old and new limits of the stricter clause, between those of the looser clause, or outside both.
\textbf{Notes.} A reader who ignores the facts achieves 0.667 item accuracy but solves only 1 of 201 contrast groups.

\subsection{\texorpdfstring{\texttt{dep\_\allowbreak{}resolve} (layer E)}{dep resolve (layer E)}}\label{sec:A.10}

\textbf{Question.} ``Does the proposed set install cleanly in the target environment?'' (\texttt{yes}, \texttt{no}).
\textbf{State.} Version-ordering conventions (e.g., 3.10.0 is later than 3.9.9), the target environment, an invented package index, and pinned package versions.
\textbf{Levels.} From 4 to 5 pins in the smallest cases to 9 to 12 in the largest, with runtime requirements and conflicts at higher levels.
\textbf{Groups.} Members differ in one pin, most often through a narrowly missed version bound (79 pairs), a missing pin (78), or an unmet requirement (78).
\textbf{Notes.} Level 1 has a median of three checks per positive item.

\subsection{\texorpdfstring{\texttt{ledger\_\allowbreak{}reconcile} (layer F)}{ledger reconcile (layer F)}}\label{sec:A.11}

\textbf{Question.} ``Which entry of ledger A does not reconcile with ledger B?''
Options are five entries from ledger A and \texttt{none}.
\textbf{State.} Two ledgers covering the same period and their matching conventions.
\textbf{Levels.} Ledgers containing 8, 14, or 22 entries, with date tolerances, split lines, fees, and batch lines at higher levels.
\textbf{Groups.} Members differ in the modified lines of ledger B, most often through a missing line (142 items) or a changed amount (140), and \texttt{none} is the gold answer for 100 items.

\subsection{\texorpdfstring{\texttt{record\_\allowbreak{}rule} (layer F)}{record rule (layer F)}}\label{sec:A.12}

\textbf{Question.} ``Which rule does the batch violate?''
Options are one key per listed rule and \texttt{none}.
\textbf{State.} A record type, 6 to 9 numbered integrity rules, reference lists, and a batch of records with at most one rule violated.
\textbf{Levels.} Batches of 10, 16, or 24 records.
\textbf{Groups.} 150 groups of four items each, including near-miss violations and, in 72 groups, the untouched batch.
Each of thirteen rule types has 40 or 41 violating items.

\subsection{\texorpdfstring{\texttt{filter\_\allowbreak{}match} (layer G)}{filter match (layer G)}}\label{sec:A.13}

\textbf{Question.} ``Does the record satisfy every constraint of the request?'' (\texttt{yes}, \texttt{no}).
\textbf{State.} A request with 6, 9, or 12 constraints of twenty types, ten of which require a small computation, and one catalog record in JSON.
\textbf{Levels.} The number of constraints.
\textbf{Groups.} In both members, the compared quantity lies at the limit or one step away, with 21 negative items per computed type and 9 for each remaining type.

\endgroup
\FloatBarrier
\section{Protocol}\label{sec:B}

\subsection{Systems}\label{sec:B.1}

\begingroup
\renewcommand{\thetable}{B1}
\footnotesize
\setlength{\tabcolsep}{5pt}
\renewcommand{\arraystretch}{1.1}

\newlength{\mdBOneWidth}
\setlength{\mdBOneWidth}{\dimexpr\linewidth-12\tabcolsep\relax}
\renewcommand{\mdw}[1]{\dimexpr#1\mdBOneWidth\relax}

\FloatBarrier
\begin{longtable}{>{\raggedright\arraybackslash}p{\mdw{0.1559}}>{\raggedleft\arraybackslash}p{\mdw{0.1418}}>{\raggedright\arraybackslash}p{\mdw{0.2412}}>{\raggedright\arraybackslash}p{\mdw{0.2425}}>{\raggedright\arraybackslash}p{\mdw{0.0926}}>{\raggedright\arraybackslash}p{\mdw{0.1130}}}

\caption{\textbf{Evaluated models, inference protocols, and source-exposure flags.} The table covers 40 decision models and 14 general models; additional reasoning settings are reported in Table~\ref{tab:D5}. Open models from other groups use the listed revisions and bf16 precision on H200 GPUs, except for Blink's FP8 build and the recommended float16 setting for d1-omni-600M. Exposure flags count named reference sources or related data families; for our models, they count tasks with at least 3\% detected training overlap before filtering. Jebadiah refers to Kev's sources without listing them, leaving nine tasks with unknown exposure status. Bracketed names indicate table abbreviations.}\label{tab:B1}\\

\toprule
\textbf{system} & \textbf{parameters} & \textbf{type} & \textbf{inference protocol} & \textbf{release status} & \textbf{exposure flags} \\
\midrule
\endfirsthead

\toprule
\textbf{system} & \textbf{parameters} & \textbf{type} & \textbf{inference protocol} & \textbf{release status} & \textbf{exposure flags} \\
\midrule
\endhead
\mdbodyrow Perplexity Decider v1.1 & 27B & decision model & open weights, revision 3b45dea, using the model-card inference protocol & released & 16 (2 unknown) \\
\mdbodyrow Kev 27B & 27B & decision model & open weights, revision af0e6d5, the authors' code at commit 5e42a7a & released & 9 \\
\mdbodyrow JEV-27B & 27B & decision model; model card identifies it as a student of the hosted reference & open weights, revision 51740a8, backbone, adapter and decision head & released & no list \\
\mdbodyrow Clef 27B & 27B & decision model & open weights, revision 2f3de3d, using the authors' serving implementation & released & no list \\
\mdbodyrow Clef flash & 9.5B & decision model & open weights, revision 17f0b0a, using the authors' serving implementation & released & no list \\
\mdbodyrow Decision 2.0 Vega 27B {[}Decision 2.0 Vega{]} & 27B & decision model, an adapter and decision head on Qwen3.8-27B & open weights, revision 7aec49a with the base at revision 1d4bf0f, using the authors' decision2 inference package & released & no list \\
\mdbodyrow Jebadiah 27B & 27B & decision model, a merged LoRA on Qwen3.8-27B & open weights, revision dfa3697, the scripts of the model repository & released & 0 (9 unknown) \\
\mdbodyrow Blink v0.3 26B-A4B {[}Blink v0.3{]} & 26B-A4B & decision model, a full fine-tune of Gemma-4-26B-A4B-it & the authors' FP8 build, revision a3b0331 (the published leaderboard uses the NVFP4 build of the same weights), the surogate decisions v1 protocol through vLLM & released & no list \\
\mdbodyrow Xor 26B-A4B 1.0 {[}Xor{]} & 26B-A4B & decision model, two merged LoRA lineages on Gemma-4-26B-A4B-it & open weights, revision 8afd439 (tag v1), the authors' serving bundle on SGLang & released & no list \\
\mdbodyrow Bespoke Nimble 9B v3 {[}Nimble v3{]} & 9B & decision model, a LoRA adapter on Qwen3.5-9B & open weights, revision 8e927b9 with the base at revision c202236, the inference engine used for the authors' Decision Index evaluation & released & no list \\
\mdbodyrow GEV-26B-Decide & 26B-A4B & decision model, a LoRA adapter and decision head on Gemma-4-26B-A4B-it & open weights, revision 7c89590, System 1 inference with the model-card Transformers configuration and reasoning disabled & released & no list \\
\mdbodyrow H2O-Lightning-4B & 4B & decision model, a fine-tune of Qwen3.5-4B & open weights, revision 457f9d3, the authors' inference wrapper, its label probabilities read through vLLM & released & no list \\
\mdbodyrow Decision 2.0 Lux 9B {[}Decision 2.0 Lux{]} & 9B & decision model, a fine-tune of Qwen3.5-9B with a decision head & open weights, revision 78bf3c0, using the authors' decision2 inference package; the 2 items longer than its 16,384-token limit unanswered & released & no list \\
\mdbodyrow Decision 2.0 Nox 4B {[}Decision 2.0 Nox{]} & 4B & decision model, a decision head on Qwen3.5-4B-Base & open weights, revision 25e8f67, using the authors' decision2 inference package; the 2 items longer than its 16,384-token limit unanswered & released & no list \\
\mdbodyrow JEV-9B & 9B & decision model; model card identifies it as a student of the hosted reference & open weights, revision b63f651, backbone, adapter and decision head & released & no list \\
\mdbodyrow d1-3B & 3B & decision model, a fine-tune of LFM2.5-VL-3B & open weights, revision 051bcc4, the inference engine provided in the model repository & released & no list \\
\mdbodyrow Decision 2.0 Sol 2B {[}Decision 2.0 Sol{]} & 2B & decision model, a fine-tune of Qwen3.5-2B with a decision head & open weights, revision 64235be, using the authors' decision2 inference package; the 2 items longer than its 16,384-token limit unanswered & released & no list \\
\mdbodyrow Hopper 12B (trained) {[}Hopper 12B{]} & 12B & decision model, a LoRA adapter on Gemma-4-12B-it & open weights, adapter revision 4b605f2 with the base at revision 707f0a3, the authors' serving code (tag 12b-1.1.0), the adapter kept unmerged & released & 4 \\
\mdbodyrow Decision 4B v1.2 (FlyMyJev-4B, preview) {[}FlyMy Decision 4B{]} & 4B & decision model, a LoRA adapter on Qwen3.5-4B & open weights, revision 79792bf with the base at revision 851bf6e, the model repository's inference implementation; 6 items unanswered (4 with more options than its 26 letters, 2 longer than its 16,384-token limit) & released & no list \\
\mdbodyrow NeoHorse-Jev-4B & 4B & decision model, a fine-tune of NeoHorse-1-4B (from Qwen3.5-4B) with a pointer head & open weights, revision 434cb21, the inference package provided in the model repository, at its default limits; the 2,627 items whose state is longer than its 2,048-token limit unanswered & released & no list \\
\mdbodyrow Hopper (G) 1.2 & 4B & decision model, a LoRA adapter on Qwen3.5-4B & open weights, adapter revision d60a1d6 with the base at revision 851bf6e, the authors' serving code (tag g-1.2.0) with its calibration map & released & 4 \\
\mdbodyrow Intern-Decision-4B & 4B & decision model, a fine-tune of Qwen3.5-4B & open weights, revision 0e5e6aa, the inference code of the model repository; the 473 items longer than its 8,192-token limit unanswered (its engine does not truncate) & released & no list \\
\mdbodyrow Intern-Decision-2B & 2B & decision model, a fine-tune of Qwen3.5-2B & open weights, revision 8797836, the inference code of the model repository; the 473 items longer than its 8,192-token limit unanswered (its engine does not truncate) & released & no list \\
\mdbodyrow this-that 1.2 & 2B & decision model, a fine-tune of decider-2b (Qwen3.5-2B lineage) & open weights, revision c4d1c30, the authors' package (GitHub commit 6af11c6), which truncates the state to its first 1,536 tokens & released & no list \\
\mdbodyrow Decision 2.0 Eos 0.8B {[}Decision 2.0 Eos{]} & 0.8B & decision model, a fine-tune of Qwen3.5-0.8B with a decision head & open weights, revision 3594047, using the authors' decision2 inference package; the 2 items longer than its 16,384-token limit unanswered & released & no list \\
\mdbodyrow d1-omni-600M & 0.6B & decision model, a fine-tune of LFM2.5-Encoder-350M with image and audio front ends & open weights, revision 02b55d7, the inference engine provided in the model repository, in float16 as recommended in the model card; states truncated to the 16,384-token context limit & released & no list \\
\mdbodyrow Decision 2.0 Kai 0.6B {[}Decision 2.0 Kai{]} & 0.6B & decision model, a fine-tune of Qwen3-0.6B-Base with a decision head & open weights, revision cd49ea3, using the authors' decision2 inference package; the 443 items longer than its 8,192-token limit unanswered & released & no list \\
\mdbodyrow Decision-4B (Eval Engine) {[}Eval Engine Decision-4B{]} & 4B & decision model, a LoRA adapter on Qwen3.5-4B & open weights, revision aadc555 with the base at revision 851bf6e, the model-card inference implementation, one prompt at a time; 3,366 items unanswered under the card's limits (3,360 prompts longer than its 2,048-token context, 6 questions with more than its 24 options) & released & 1 \\
\mdbodyrow CoCo-Decision-4B-Ko & 4B & decision model, a LoRA adapter on Qwen3.5-4B & open weights, revision 24aea63 (tag v1.0.0) with the base at revision 851bf6e, the model-card inference implementation; the 4 items with more options than its 26 letters unanswered & released & no list \\
\mdbodyrow strands-decider-2B-hobson-v21 {[}Strands decider 2B v21{]} & 2B & decision model, a LoRA adapter and pointer head on Qwen3.5-2B-Base & open weights, revision 2b52a62 with the base at revision b1485b2, the authors' package (GitHub commit 0ac22a9), which truncates the state to the checkpoint's 4,096-token window & released & 3 \\
\mdbodyrow Bosun 3.1 1.7B & 1.7B & decision model, a LoRA adapter and trained decision tokens on Qwen3-1.7B & open weights, revision 1d8dc82 with the base at revision 70d244c, the code of the model repository (its predict function, candidates shuffled by the loader into fixed decision slots) & released & no list \\
\mdbodyrow Decision 3.0 d3 & 27B & decision model, on Qwen3.8-27B with a decision head and a vision encoder & open weights, revision cbff3a3, the authors' runtime in the model repository (\texttt{D3.system\_one}), no input limit & released & no list \\
\mdbodyrow Decision 3.0 d3-flash & 9B & decision model, on Qwen3.5-9B with a decision head and a vision encoder & open weights, revision 521525e, the authors' runtime in the model repository (\texttt{D3.system\_one}), no input limit & released & no list \\
\mdbodyrow Decision 3.0 d3-mini & 4B & decision model, on Qwen3.5-4B with a decision head and a vision encoder & open weights, revision a475367, the authors' runtime in the model repository (\texttt{D3.system\_one}), no input limit & released & no list \\
\mdbodyrow Decision 3.0 d3-nano & 2B & decision model, on Qwen3.5-2B with a decision head and a vision encoder & open weights, revision 3aeb1c2, the authors' runtime in the model repository (\texttt{D3.system\_one}), no input limit & released & no list \\
\mdbodyrow Decision 3.0 d3-lite & 0.8B & decision model, on Qwen3.5-0.8B with a decision head and a vision encoder & open weights, revision a8d25a9, the authors' runtime in the model repository (\texttt{D3.system\_one}), no input limit & released & no list \\
\mdbodyrow Decision 3.0 d3-edge & 0.6B & decision model, 12 of the 24 text layers of Qwen3.5-0.8B with a decision head and a vision encoder & open weights, revision 18d61df, the authors' runtime in the model repository (\texttt{D3.system\_one}), no input limit & released & no list \\
\mdbodyrow hosted reference (jev-1.13.0) & -- & hosted decision model & API, called on 7 and 9 October 2026 & API only & no list \\
\mdbodyrow DeepSeek-V4-Pro & -- & general chat model & API, no reasoning, all 7,800 core items & API only & unknown \\
\mdbodyrow GLM-5.3 & -- & general chat model & API, lowest effort on all 7,800 core items, reasoning on the sample of 1,296 & API only & unknown \\
\mdbodyrow DeepSeek-V4.1-Flash & -- & general chat model & API, no reasoning, all 7,800 core items & API only & unknown \\
\mdbodyrow Qwen3.8-27B & 27B & general open model & open weights, revision 1d4bf0f, one pass, probabilities of the option keys, thinking off & released & unknown \\
\mdbodyrow Qwen3.6-35B-A3B & 35B-A3B & general open model & open weights, revision 995ad96, one pass, probabilities of the option keys, thinking off & released & unknown \\
\mdbodyrow Gemma-4-31B-it & 31B & general open model & open weights, revision 842da37, one pass, probabilities of the option keys, thinking off & released & unknown \\
\mdbodyrow Gemma-4-26B-A4B-it & 26B-A4B & general open model & open weights, revision 4d7ae49, one pass, probabilities of the option keys, thinking off & released & unknown \\
\mdbodyrow Phi-4 & 14B & general open model & open weights, revision 2db69c1, one pass, probabilities of the option keys; the 12 items longer than its 16,384-token context unanswered & released & unknown \\
\mdbodyrow Gemma-3-12B-it & 12B & general open model & the Unsloth upload of the weights, revision 9478e66, one pass, probabilities of the option keys & released & unknown \\
\mdbodyrow Qwen3.5-9B & 9B & general open model & open weights, revision c202236, one pass, probabilities of the option keys, thinking off & released & unknown \\
\mdbodyrow Granite-3.3-8B-Instruct & 8B & general open model & open weights, revision 51dd4bc, one pass, probabilities of the option keys & released & unknown \\
\mdbodyrow Qwen3.5-4B & 4B & general open model & open weights, revision 851bf6e, one pass, probabilities of the option keys, thinking off & released & unknown \\
\mdbodyrow Qwen3.5-2B & 2B & general open model & open weights, revision 15852e8, one pass, probabilities of the option keys, thinking off & released & unknown \\
\mdbodyrow Qwen3.5-0.8B & 0.8B & general open model & open weights, revision 2fc0636, one pass, probabilities of the option keys, thinking off & released & unknown \\
\mdourrow StartLux-Decision-27B-Instant {[}SL-27B Instant{]} & 27B & decision model (ours) & one pass with bidirectional prompt processing & released & 16 \\
\mdourrow StartLux-Decision-27B-AutoThink {[}SL-27B AutoThink{]} & 27B & decision model (ours) & one pass, with a reasoning trace of at most 4,096 tokens when the top probability is below 0.9 & released & 16 \\
\bottomrule
\end{longtable}
\endgroup

GLM-5.3 does not support disabling reasoning; even its lowest-effort setting uses a hidden reasoning trace, averaging 121 tokens per item on the subset in Table~\ref{tab:D5}.
The three API chat models report a top probability of at least 0.99 in 63\% to 93\% of their responses.
We report their accuracy and pair accuracy as reference points, without treating stated probabilities as directly comparable to decision-head or option-key probabilities.
Confidence metrics for open general models are reported separately in Table~\ref{tab:Dgen} and depend on the probability-readout protocol.
% TODO: document each open-model prompt, option tokenization, multi-token scoring, normalization, and treatment of probability mass outside the allowed options.

\subsection{Requests and answer files}\label{sec:B.2}

\textbf{Answer files.}
Each system has one answer file per task at
\texttt{runs/\allowbreak{}\textless{}system\textgreater{}/\allowbreak{}\textless{}task\textgreater{}.jsonl}.
A representative record is
\texttt{\{\textquotedbl{}id\textquotedbl{}:
\textquotedbl{}policy\_\allowbreak{}case:00007\textquotedbl{}, \textquotedbl{}p\textquotedbl{}:
\{\textquotedbl{}approve\textquotedbl{}: 0.12, \textquotedbl{}deny\textquotedbl{}: 0.80,
\textquotedbl{}escalate\textquotedbl{}: 0.08\}\}}.
The runners store the SHA-256 hash of the evaluated items file alongside the answers.
Reasoning runs may also record \texttt{tokens} and \texttt{closed}; the latter indicates whether the model ended its reasoning trace itself.

Task inclusion requires records (answers or explicit errors) for at least 99\% of items; this is distinct from successful-prediction coverage.
Failed or missing predictions count as incorrect and use a uniform option distribution for confidence-based scoring.
The scorer normalizes valid probabilities over the listed options and breaks top-option ties in the listed order.

\subsection{Prompt format for chat models}\label{sec:B.3}

General chat models receive the state under \texttt{STATE}, the instruction under \texttt{QUESTION}, and the options as \texttt{key: meaning} lines under \texttt{OPTIONS (key: meaning)}.
Structured states are formatted as indented JSON.
The system message is reproduced below.
The user message ends with ``Reply with the JSON object, in this form:'', followed by an object containing the item's option keys.

\begin{lstlisting}
You answer one typed question about the state you are given. Reply with the key of exactly one
option, copied as it is written, and a probability for every option, as one JSON object and nothing
else: {"choice": "<the key of the option you pick>", "probabilities": {"<key>": <probability>, ...}}.
"probabilities" has one number between 0 and 1 for every option key, the numbers add up to 1, and
each number is how likely you think it is that this option is the right one.
\end{lstlisting}

The predicted option is the one with the highest stated probability.
At the lower-effort settings, requests used temperature 0 and output budgets of 2,048 tokens for DeepSeek-V4-Pro with reasoning disabled, or 8,192 tokens for GLM-5.3 at its lowest reasoning effort.
Higher-effort runs used the provider's default sampling, high effort, and an output budget of 32,768 tokens.
Replies for the three families unchanged since version 1.0 were reused, and all 18,192 replies underlying version 1.1 were parsed successfully.
% Internal parsing provenance (not rendered): v1_1/general_llm/run_log.txt.

\subsection{Worked examples and intervals}\label{sec:B.4}

\textbf{Worked examples.}
If every item has a top probability of 0.90 and 70\% of predictions are correct, the calibration error is 20.0.
If 5 of 10 items have a top probability of 0.90 and 4 of these are correct, accuracy at 0.90 is 80.0 with a share of 50.0.

\textbf{Intervals.}
Bootstrap resampling uses groups for grouped tasks and individual items otherwise, with matched resamples for paired comparisons and seed 20261007.
Reranking tasks resample queries, keeping each query's candidates together.
Each resample recomputes task metrics, chance correction, zero flooring, layer averages, and the reference total; core metrics retain their stated family-macro or pooled-item aggregation.
Analysis-specific resample counts are reported in the corresponding captions.
Calibration error uses 15 equal-width confidence bins and is reported as a point estimate; its sampling uncertainty is not quantified.
\href{https://github.com/StartLuxLabs/DecisionPivot/blob/2b7000f312ccbbdce4e8625b56d8a8186ea0047a/tests/test_score.py}{Unit tests} for the scoring implementation are provided in the benchmark code repository.
\FloatBarrier
\section{Reference tasks and exposure}\label{sec:C}

Table~\ref{tab:C1} summarizes coverage across the seven decision layers, and Table~\ref{tab:C2} details all 35 converted reference tasks.
In Table~\ref{tab:C2}, Rule denotes the best content-blind baseline for accuracy-based tasks, while StartLux reports the proportion of items detected in our training data before filtering; tasks with at least 3\% overlap are flagged as exposed.
For Perplexity and Kev, \emph{own} indicates that the training list names the dataset, \emph{family} indicates a related dataset, -- indicates neither, and ? denotes unavailable or inconclusive documentation.
Release identifies the stated item-release route (I), share-alike route (SA), or availability of reconstruction scripts (R).
Reconstruction support does not itself establish permission to redistribute converted items.
Table~\ref{tab:C4} records source-specific terms and qualifications.

Perplexity Decider v1.1 lists 13 procedural families in its training manifest, including arithmetic, policy applicability, and table lookup.
Kev 27B describes generated policy and rule records and seven generated families without releasing the generators.
JEV-27B describes 498,010 synthetic scenarios labeled with the hosted reference's output distributions, alongside 94,801 programmatic rows.

\begin{table}[!htbp]
\renewcommand{\thetable}{C1}
\centering
\caption{\textbf{Reference-track coverage across seven decision layers.} The track contains 32,233 items from 35 tasks, including 21 with human labels. Labels gives the numbers of tasks with human, programmatic, or mixed labels. Exposed counts tasks with at least 3\% detected item overlap before filtering for our models, and tasks whose sources or related data families are named in public training documentation for Perplexity Decider v1.1 and Kev 27B.}\label{tab:C1}
{\footnotesize\setlength{\tabcolsep}{5pt}\renewcommand{\arraystretch}{1.1}
\noindent\begin{adjustbox}{max width=\linewidth}
\begin{tabular}{>{\raggedright\arraybackslash}p{\mdw{0.1563}}>{\raggedleft\arraybackslash}p{\mdw{0.0669}}>{\raggedleft\arraybackslash}p{\mdw{0.0712}}>{\raggedright\arraybackslash}p{\mdw{0.4050}}>{\raggedleft\arraybackslash}p{\mdw{0.0816}}>{\raggedleft\arraybackslash}p{\mdw{0.2058}}}
\toprule
\textbf{layer} & \textbf{tasks} & \textbf{items} & \textbf{two of its decisions} & \textbf{labels} & \textbf{exposed: StartLux / Perplexity / Kev} \\
\midrule
\mdbodyrow A before generation & 6 & 5,112 & is the question ambiguous / will a named model answer it correctly & 4 / 2 / 0 & 2 / 2 / 4 \\
\mdbodyrow B guards & 7 & 3,563 & is the prompt unsafe / does the web page carry an injected instruction & 4 / 2 / 1 & 3 / 4 / 1 \\
\mdbodyrow C after generation & 8 & 5,845 & which response is better / which step of a solution is the first wrong one & 4 / 1 / 3 & 7 / 4 / 1 \\
\mdbodyrow D agent loop & 3 & 1,597 & which page element to act on next / did the web agent complete its task & 2 / 0 / 1 & 2 / 0 / 0 \\
\mdbodyrow E reasoning classification & 4 & 2,301 & does the contract hold a clause of this kind / is the conclusion true, false or undetermined & 2 / 1 / 1 & 2 / 3 / 2 \\
\mdbodyrow F data pipelines & 1 & 996 & is the label that the data set gives this text correct & 1 / 0 / 0 & 0 / 1 / 1 \\
\mdbodyrow G search and matching & 6 & 12,819 & do the two records describe the same product / which of two documents that differ by a negation is relevant & 4 / 0 / 2 & 0 / 2 / 0 \\
\mdbodyrow all & 35 & 32,233 &  & 21 / 6 / 8 & 16 / 16 / 9 \\
\bottomrule
\end{tabular}
\end{adjustbox}\par}
\end{table}

\begingroup
\renewcommand{\thetable}{C2}
\scriptsize\setlength{\tabcolsep}{4pt}\renewcommand{\arraystretch}{1.1}
\FloatBarrier
\begin{longtable}{>{\raggedright\arraybackslash}p{\mdw{0.1443}}>{\raggedright\arraybackslash}p{\mdw{0.1884}}>{\raggedleft\arraybackslash}p{\mdw{0.0593}}>{\raggedright\arraybackslash}p{\mdw{0.1132}}>{\raggedright\arraybackslash}p{\mdw{0.1049}}>{\raggedleft\arraybackslash}p{\mdw{0.0529}}>{\raggedleft\arraybackslash}p{\mdw{0.0967}}>{\raggedright\arraybackslash}p{\mdw{0.0987}}>{\raggedright\arraybackslash}p{\mdw{0.0599}}>{\raggedright\arraybackslash}p{\mdw{0.0694}}}
\caption{\textbf{Reference tasks retain their source labels and evaluation metrics.} AUROC denotes area under the receiver operating characteristic curve; nDCG@10 denotes normalized discounted cumulative gain at rank 10.}\label{tab:C2}\\
\toprule
\textbf{task} & \textbf{decision} & \textbf{items} & \textbf{labels} & \textbf{metric (chance)} & \textbf{rule} & \textbf{StartLux} & \textbf{Perplexity} & \textbf{Kev} & \textbf{release} \\
\midrule
\endfirsthead
\multicolumn{10}{l}{\textit{Table~\thetable{} (continued)}}\\
\toprule
\textbf{task} & \textbf{decision} & \textbf{items} & \textbf{labels} & \textbf{metric (chance)} & \textbf{rule} & \textbf{StartLux} & \textbf{Perplexity} & \textbf{Kev} & \textbf{release} \\
\midrule
\endhead
\bottomrule
\endfoot
\mdbodyrow A \texttt{ambigqa} & does the question have several readings & 990 & human & accuracy (0.500) & 0.554 & -- (0.8\%) & -- & -- & SA \\
\mdbodyrow A \texttt{cap\_\allowbreak{}boundary\_\allowbreak{}card} & will the named chat model answer correctly & 911 & programmatic & skill (0.000) & -- & exposed (8.0\%) & family & family & R \\
\mdbodyrow A \texttt{cap\_\allowbreak{}boundary\_\allowbreak{}record} & the same, with the model's record on related questions & 774 & programmatic & skill (0.000) & -- & exposed (21.8\%) & family & family & R \\
\mdbodyrow A \texttt{freshqa\_\allowbreak{}premise} & does the question rest on a false premise & 495 & human & AUROC (0.500) & -- & -- (1.0\%) & -- & -- & R \\
\mdbodyrow A \texttt{known\_\allowbreak{}unknown} & does the question have a known answer & 972 & human & accuracy (0.500) & 0.580 & -- (1.7\%) & -- & family & I \\
\mdbodyrow A \texttt{selfaware} & can the question be answered at all & 970 & human & accuracy (0.500) & 0.571 & -- (1.6\%) & -- & family & SA \\
\mdbodyrow B \texttt{aegis2\_\allowbreak{}prompt} & is the prompt unsafe & 836 & human & accuracy (0.500) & 0.584 & exposed (13.7\%) & family & own & I \\
\mdbodyrow B \texttt{bipia\_\allowbreak{}injection} & does the e-mail carry an inserted instruction & 88 & programmatic & accuracy (0.500) & 0.602 & -- (0.0\%) & family & -- & I \\
\mdbodyrow B \texttt{browsesafe} & does the web page carry an injected instruction & 1,000 & programmatic & accuracy (0.500) & 0.506 & -- (0.0\%) & ? & -- & I \\
\mdbodyrow B \texttt{cosapien\_\allowbreak{}policy} & allowed, disallowed or partly allowed under a safety configuration & 148 & human & accuracy (0.333) & 0.432 & -- (0.0\%) & -- & -- & I \\
\mdbodyrow B \texttt{jailbreak\_\allowbreak{}wild} & is the prompt a jailbreak attempt & 796 & mixed & accuracy (0.500) & 0.528 & exposed (6.5\%) & family & -- & I \\
\mdbodyrow B \texttt{jbb\_judge} & does the response give harmful content & 250 & human & AUROC (0.500) & -- & exposed (16.7\%) & -- & -- & I \\
\mdbodyrow B \texttt{xstest} & should the request be refused & 445 & human & AUROC (0.500) & -- & -- (1.1\%) & family & -- & I \\
\mdbodyrow C \texttt{contextual\_\allowbreak{}judge} & which response is better for the context & 877 & mixed & accuracy (0.500) & 0.502 & exposed (12.1\%) & -- & -- & I \\
\mdbodyrow C \texttt{faithbench} & is the summary faithful to its source & 298 & human & accuracy (0.500) & 0.550 & exposed (5.5\%) & -- & -- & R \\
\mdbodyrow C \texttt{judgebench} & which response is correct & 619 & programmatic & accuracy (0.500) & 0.549 & -- (0.0\%) & -- & -- & I \\
\mdbodyrow C \texttt{llmbar} & which output follows the instruction better & 355 & human & accuracy (0.500) & 0.730 & exposed (14.8\%) & family & -- & I \\
\mdbodyrow C \texttt{processbench} & which step is the first wrong one, or none & 939 & human & accuracy (0.142) & 0.366 & exposed (3.8\%) & family & family & I \\
\mdbodyrow C \texttt{rewardbench} & which response is better & 888 & mixed & accuracy (0.500) & 0.561 & exposed (10.9\%) & family & -- & R \\
\mdbodyrow C \texttt{rewardbench2} & which of four responses is best & 953 & mixed & accuracy (0.250) & 0.346 & exposed (4.7\%) & family & -- & I \\
\mdbodyrow C \texttt{squad2\_\allowbreak{}answerable} & can the question be answered from the passage & 916 & human & accuracy (0.500) & 0.546 & exposed (5.1\%) & ? & -- & SA \\
\mdbodyrow D \texttt{agentreward\_\allowbreak{}success} & did the web agent complete its task & 552 & human & accuracy (0.500) & 0.549 & -- (0.2\%) & -- & -- & R \\
\mdbodyrow D \texttt{bfcl\_\allowbreak{}relevance} & does any listed function serve the request & 144 & mixed & accuracy (0.500) & 0.562 & exposed (35.3\%) & -- & -- & I \\
\mdbodyrow D \texttt{mind2web\_\allowbreak{}element} & which of ten page elements to act on next & 901 & human & accuracy (0.100) & 0.114 & exposed (9.9\%) & -- & -- & R \\
\mdbodyrow E \texttt{cuad\_clause} & does the contract hold a clause of the named category & 192 & human & accuracy (0.500) & 0.562 & exposed (94.1\%) & family & family & I \\
\mdbodyrow E \texttt{folio\_\allowbreak{}validation} & is the conclusion true, false or undetermined & 203 & human & accuracy (0.333) & 0.355 & -- (0.5\%) & own & -- & SA \\
\mdbodyrow E \texttt{legalbench\_\allowbreak{}mix} & ten legal classification sub-tasks & 910 & mixed & accuracy (0.429) & 0.455 & exposed (6.6\%) & family & family & I \\
\mdbodyrow E \texttt{primevul\_\allowbreak{}pair} & is the function vulnerable (asked for a function and its patch) & 996 & programmatic & pair accuracy (0.250) & -- & -- (0.4\%) & -- & -- & I \\
\mdbodyrow F \texttt{label\_\allowbreak{}errors} & is the label of the text correct & 996 & human & accuracy (0.500) & 0.524 & -- (2.0\%) & family & own & R \\
\mdbodyrow G \texttt{entity\_\allowbreak{}match\_\allowbreak{}abt\_\allowbreak{}buy} & do the two records describe the same product & 996 & mixed & AUROC (0.500) & -- & -- (0.4\%) & -- & -- & R \\
\mdbodyrow G \texttt{entity\_\allowbreak{}match\_\allowbreak{}amazon\_\allowbreak{}google} & do the two records describe the same product & 999 & mixed & AUROC (0.500) & -- & -- (0.1\%) & -- & -- & R \\
\mdbodyrow G \texttt{nevir} & which of two documents that differ by a negation is relevant & 998 & human & pair accuracy (0.250) & -- & -- (0.2\%) & family & -- & I \\
\mdbodyrow G \texttt{paws} & do the two sentences mean the same & 976 & human & accuracy (0.500) & 0.537 & -- (1.5\%) & -- & -- & R \\
\mdbodyrow G \texttt{rerank\_\allowbreak{}nfcorpus} & how well does the passage answer the query & 4,950 & human & nDCG@10 (0.142) & -- & -- (1.0\%) & -- & -- & R \\
\mdbodyrow G \texttt{rerank\_\allowbreak{}scifact} & how well does the passage answer the query & 3,900 & human & nDCG@10 (0.088) & -- & -- (0.5\%) & family & -- & R \\
\end{longtable}
\endgroup

\textbf{Training-source flags do not establish item exposure or its effects.}
Table~\ref{tab:C3} compares performance on tasks whose sources are named or not named in available training documentation.
After adjusting for the corresponding gap among four selected control models, only Kev 27B has an interval excluding zero.
For our final models, after filtering detected item overlap, the adjusted gaps are +0.6 [$-$4.3, +6.0] for Instant and +1.7 [$-$3.1, +7.1] for AutoThink.
These intervals do not rule out training effects.
The comparison remains observational: source naming may correlate with task difficulty or model specialization, and the controls are not known to be unexposed.
Conversely, strong performance on a generated task does not establish prior exposure; Perplexity Decider performs well on \texttt{dep\_resolve} although this task type is absent from its published list.

\begin{table}[!htbp]
\renewcommand{\thetable}{C3}
\centering
\caption{\textbf{Observational comparison by training-source flags.} Named tasks have a source or related family listed in training documentation; for our models, the flag denotes at least 3\% detected item overlap before filtering. Gap is the difference between mean floored, chance-corrected task scores on named and unlisted sources, with 95\% intervals from 1,000 task resamples. Adjusted gap subtracts the corresponding gap for four selected controls: the hosted model, Clef 27B, Clef flash, and JEV-27B. Their exposure is unknown, so adjustment does not identify a causal training effect. Two Perplexity tasks are unknown.}\label{tab:C3}
% Internal analysis provenance (not rendered): Source: \texttt{v1\_\allowbreak{}3/\allowbreak{}analysis/\allowbreak{}out/\allowbreak{}t7\_\allowbreak{}exposure.csv}; our overlap scan is dated 10 October 2026.
{\footnotesize\setlength{\tabcolsep}{4pt}\renewcommand{\arraystretch}{1.1}
\noindent\begin{adjustbox}{max width=\linewidth}
\begin{tabular}{llrrrrrr}
\toprule
\textbf{system} & \textbf{list} & {\begin{tabular}[b]{@{}r@{}}\textbf{tasks}\\\textbf{named /}\\\textbf{not named}\end{tabular}} & {\begin{tabular}[b]{@{}r@{}}\textbf{mean}\\\textbf{named}\end{tabular}} & {\begin{tabular}[b]{@{}r@{}}\textbf{mean}\\\textbf{not}\\\textbf{named}\end{tabular}} & \textbf{gap} & {\begin{tabular}[b]{@{}r@{}}\textbf{gap of}\\\textbf{the}\\\textbf{controls}\end{tabular}} & \textbf{adjusted gap} \\
\midrule
\mdbodyrow Perplexity Decider v1.1 27B & manifest & 16 / 17 & 63.7 & 58.7 & +5.1 {[}$-$15.0, +27.1{]} & +3.5 & +1.6 {[}$-$4.0, +7.8{]} \\
\mdbodyrow Kev 27B & model card & 9 / 26 & 47.7 & 61.0 & $-$13.2 {[}$-$30.8, +3.5{]} & $-$18.7 & +5.4 {[}+0.5, +11.0{]} \\
\mdourrow SL-27B Instant & our data & 16 / 19 & 62.8 & 64.8 & $-$2.0 {[}$-$20.4, +17.4{]} & $-$2.6 & +0.6 {[}$-$4.3, +6.0{]} \\
\mdourrow SL-27B AutoThink & our data & 16 / 19 & 62.9 & 63.8 & $-$0.9 {[}$-$19.1, +18.1{]} & $-$2.6 & +1.7 {[}$-$3.1, +7.1{]} \\
\bottomrule
\end{tabular}
\end{adjustbox}\par}
\end{table}

\begingroup
\renewcommand{\thetable}{C4}
\scriptsize\setlength{\tabcolsep}{4pt}\renewcommand{\arraystretch}{1.1}
\FloatBarrier
\begin{longtable}{>{\raggedright\arraybackslash}p{\mdw{0.22}}>{\raggedright\arraybackslash}p{\mdw{0.28}}>{\raggedright\arraybackslash}p{\mdw{0.28}}>{\raggedright\arraybackslash}p{\mdw{0.22}}}
\caption{\textbf{Recorded source terms for the 35 reference tasks.} Licence statements follow source cards or license files, with unresolved or source-specific qualifications shown explicitly. I and SA denote the stated item-release routes; R denotes reconstruction-script availability, not an independent statement of redistribution status. Sampling targets at most 1,000 items per task, or 100 queries with 50 candidates for reranking. Final counts follow the filtering described in Section~\ref{sec:quality}. Core items are generated without third-party text and use CC BY 4.0.}\label{tab:C4}\\
\toprule
\textbf{task} & \textbf{source} & \textbf{license as stated} & \textbf{use} \\
\midrule
\endfirsthead
\multicolumn{4}{l}{\textit{Table~\thetable{} (continued)}}\\
\toprule
\textbf{task} & \textbf{source} & \textbf{license as stated} & \textbf{use} \\
\midrule
\endhead
\bottomrule
\endfoot
A \texttt{ambigqa} & AmbigQA (AmbigNQ) & CC BY-SA 3.0 & SA \\
A \texttt{cap\_\allowbreak{}boundary\_\allowbreak{}card} & questions of NQ, PopQA, WebQuestions, TriviaQA, BBH, StrategyQA, FOLIO, ProofWriter, DROP, SVAMP, ASDiv, GSM-Hard, GSM-Plus and MAWPS & labels ours, question texts under the licenses of their sources & R, labels from our sampled answers of Gemma 3 4B and 12B, Phi-4-mini and Phi-4 \\
A \texttt{cap\_\allowbreak{}boundary\_\allowbreak{}record} & as above, with the model's record on related questions & as above & R, as above \\
A \texttt{freshqa\_\allowbreak{}premise} & FreshQA, sheet of 21 April 2026 & Apache-2.0 (repository), no statement on the question sheet & R \\
A \texttt{known\_\allowbreak{}unknown} & Known-Unknown Questions & MIT & I \\
A \texttt{selfaware} & SelfAware & CC BY-SA 4.0 & SA \\
B \texttt{aegis2\_\allowbreak{}prompt} & Aegis 2.0 (Nemotron Content Safety V2) & CC BY 4.0 & I \\
B \texttt{bipia\_\allowbreak{}injection} & BIPIA & MIT & I \\
B \texttt{browsesafe} & BrowseSafe-Bench & MIT & I \\
B \texttt{cosapien\_\allowbreak{}policy} & CoSApien & CDLA-Permissive-2.0 & I \\
B \texttt{jailbreak\_\allowbreak{}wild} & In-the-Wild Jailbreak Prompts & MIT & I \\
B \texttt{jbb\_\allowbreak{}judge} & JailbreakBench, judge comparison & MIT & I \\
B \texttt{xstest} & XSTest & CC BY 4.0 & I \\
C \texttt{contextual\_\allowbreak{}judge} & ContextualJudgeBench & CC BY 4.0 & I \\
C \texttt{faithbench} & FaithBench & CC BY-NC-SA 4.0 & R \\
C \texttt{judgebench} & JudgeBench & MIT & I \\
C \texttt{llmbar} & LLMBar & MIT & I \\
C \texttt{processbench} & ProcessBench & Apache-2.0 & I \\
C \texttt{rewardbench} & RewardBench, filtered split & ODC-BY, parts under their own licenses, some non-commercial & R \\
C \texttt{rewardbench2} & RewardBench 2 & ODC-BY, for research and educational use & I \\
C \texttt{squad2\_\allowbreak{}answerable} & SQuAD 2.0 & CC BY-SA 4.0 & SA \\
D \texttt{agentreward\_\allowbreak{}success} & AgentRewardBench & no license tag, research use under the terms of the card & R \\
D \texttt{bfcl\_\allowbreak{}relevance} & Berkeley Function Calling Leaderboard & Apache-2.0 & I \\
D \texttt{mind2web\_\allowbreak{}element} & Multimodal-Mind2Web, text columns & OpenRAIL tag (the Mind2Web card says CC BY 4.0) & R \\
E \texttt{cuad\_\allowbreak{}clause} & CUAD v1 & CC BY 4.0 & I \\
E \texttt{folio\_\allowbreak{}validation} & FOLIO & CC BY-SA 4.0 (repository), MIT tag on the gated card & SA \\
E \texttt{legalbench\_\allowbreak{}mix} & LegalBench, ten sub-tasks & CC BY 4.0 (seven sub-tasks), MIT (three) & I \\
E \texttt{primevul\_\allowbreak{}pair} & PrimeVul & MIT & I \\
F \texttt{label\_\allowbreak{}errors} & cleanlab label-errors (IMDB, Amazon reviews, 20 Newsgroups) & Apache-2.0 (annotations), texts under the terms of their sources, not stated & R \\
G \texttt{entity\_\allowbreak{}match\_\allowbreak{}abt\_\allowbreak{}buy} & Abt-Buy, DeepMatcher version & none stated by DeepMatcher, CC BY 4.0 on the page of the original records & R \\
G \texttt{entity\_\allowbreak{}match\_\allowbreak{}amazon\_\allowbreak{}google} & Amazon-Google, DeepMatcher version & as above & R \\
G \texttt{nevir} & NevIR & MIT & I \\
G \texttt{paws} & PAWS & free use, acknowledgement of Google requested & R \\
G \texttt{rerank\_\allowbreak{}nfcorpus} & system1-rerank-bench, BEIR NFCorpus & MIT (candidate lists), texts and judgments not confirmed & R \\
G \texttt{rerank\_\allowbreak{}scifact} & system1-rerank-bench, BEIR SciFact & MIT (candidate lists), CC BY-NC 2.0 (texts and judgments, as listed by BEIR) & R \\
\end{longtable}
\endgroup

\FloatBarrier
\subsection{Provenance and release routes}

The 7,800 core items contain generated text and invented entities, with no third-party text, and are released under CC BY 4.0; the code uses Apache-2.0.
The 32,233 reference items remain subject to their original copyright and source-specific terms.
Task cards document sources, versions, and licenses (Tables~\ref{tab:license} and \ref{tab:C4}).
The 35 reference sources include twenty with permissive licenses, four with share-alike licenses, two with non-commercial licenses, two with OpenRAIL or research-use terms, and seven with partly unstated or unconfirmed text licenses.
The two capability-boundary tasks pair public questions with labels derived from sampled answers of four public chat models; the sampled answers are not released.

The stated release includes generated core items, converted reference data, and reconstruction scripts, with dataset access available on request through Hugging Face.
For 13 reference tasks, scripts reconstruct items from official source files and verify their counts and hashes, including tasks with restrictive or partly unstated terms.
Reconstruction support and redistribution permission are distinct; Tables~\ref{tab:license} and \ref{tab:C4} document the recorded source terms and release routes.
The stated access conditions limit the items to evaluation and prohibit model training.
Layer B includes harmful text as classification input, while core items use invented entities and contain no personal data.
Appendix~\ref{sec:F} provides the datasheet \citep{gebru2021datasheets}.
% TODO: verify the release manifest and align all inclusion claims and access-condition descriptions with the stated source terms; earlier drafts disagree on whether the 13 reconstructable reference tasks are also redistributed.
% TODO: sources excluded for license reasons (none recorded in the paper or in external/bench; state "none" only if that is confirmed).
% TODO: terms of use of the four chat models whose sampled answers label cap_boundary_card / cap_boundary_record (Gemma 3 4B, Gemma 3 12B, Phi-4-mini, Phi-4), and the licenses of their question sources (task cards still say "to be listed per source").
% TODO: confirm that the non-commercial and research-only sources (faithbench, rerank_scifact, agentreward_success, mind2web_element) may be redistributed in the gated data set, and that the terms of the hosted reference allow its answer files to be released.

\begin{table}[!htbp]
\renewcommand{\thetable}{C5}
\centering
\caption{\textbf{Source terms and reconstruction support.} The table groups the 35 reference sources by their recorded terms and lists the generated core track separately. I denotes the stated item-release route, SA the share-alike route, and R the availability of a reconstruction script. R alone does not specify redistribution status. Table~\ref{tab:C4} lists individual sources and qualifications.}\label{tab:license}
{\footnotesize\setlength{\tabcolsep}{5pt}\renewcommand{\arraystretch}{1.1}
\noindent\begin{adjustbox}{max width=\linewidth}
\begin{tabular}{>{\raggedright\arraybackslash}p{\mdw{0.17}}>{\raggedright\arraybackslash}p{\mdw{0.24}}>{\raggedleft\arraybackslash}p{\mdw{0.06}}>{\raggedright\arraybackslash}p{\mdw{0.43}}>{\raggedright\arraybackslash}p{\mdw{0.10}}}
\toprule
\textbf{terms} & \textbf{licenses as stated} & \textbf{tasks} & \textbf{tasks or families} & \textbf{release} \\
\midrule
\mdourrow \textbf{Generated by us} & CC BY 4.0 (items), Apache-2.0 (code) & 13 & the 13 core families of Table~\ref{tab:2} & I \\
\midrule
\mdgrouprow \multicolumn{5}{l}{\textbf{Converted public test sets}} \\
Permissive & CC BY 4.0, MIT, Apache-2.0, CDLA-Permissive-2.0, ODC-BY, a free-use notice & 20 & aegis2\_prompt, xstest, contextual\_judge, cuad\_clause, legalbench\_mix, known\_unknown, bipia\_injection, browsesafe, jailbreak\_wild, jbb\_judge, judgebench, llmbar, nevir, primevul\_pair, processbench, bfcl\_relevance, cosapien\_policy, rewardbench2 (I), rewardbench and paws (R) & I, R \\
Share-alike & CC BY-SA 3.0, CC BY-SA 4.0 & 4 & ambigqa, selfaware, squad2\_answerable, folio\_validation & SA \\
Non-commercial & CC BY-NC-SA 4.0, CC BY-NC 2.0 (texts) & 2 & faithbench, rerank\_scifact & R \\
Use terms only & OpenRAIL tag, research-use terms of use & 2 & mind2web\_element, agentreward\_success & R \\
Partly unstated & no license stated or confirmed for part of the text & 7 & freshqa\_premise, label\_errors, entity\_match\_abt\_buy, entity\_match\_amazon\_google, rerank\_nfcorpus, cap\_boundary\_card, cap\_boundary\_record & R \\
\bottomrule
\end{tabular}
\end{adjustbox}\par}
\end{table}

\FloatBarrier
\section{Full results}\label{sec:D}

This appendix presents detailed results for core-track version 1.1 and reference-track version 1.3, together with the version-1.0 input-sensitivity analyses identified below.
% Internal result index (not rendered): external/bench/v1_1/report.md.
Tables~\ref{tab:D1}--\ref{tab:D4} present confidence metrics and results by task family, decision layer, and reference task.
Table~\ref{tab:D5} compares inference configurations on a core subset; Tables~\ref{tab:D6} and \ref{tab:D7} report diagnostic analyses, while Tables~\ref{tab:D8} and \ref{tab:D9} examine input sensitivity.
Model names follow the abbreviations in Table~\ref{tab:4}.

\begin{table}[!htbp]
\renewcommand{\thetable}{D1}
\centering
\caption{\textbf{Confidence-based metrics on the core track.} Values are percentages, with 95\% intervals where available. Auto at 5\% is retrospective empirical coverage macro-averaged across 13 families; accuracy and coverage at 0.90 are computed on pooled items. Calibration error uses 15 bins and is reported without an interval (Appendix~\ref{sec:B.4}). API chat models are excluded because their stated probabilities use a different readout protocol.}\label{tab:D1}
{\small\setlength{\tabcolsep}{6pt}\renewcommand{\arraystretch}{1.1}
\noindent\begin{adjustbox}{max width=\linewidth}
\begin{tabular}{lrrrr}
\toprule
\textbf{system} & \textbf{auto at 5\%} & \textbf{acc. at 0.90} & \textbf{share at 0.90} & {\begin{tabular}[b]{@{}r@{}}\textbf{calibration}\\\textbf{error}\end{tabular}} \\
\midrule
\mdbodyrow Perplexity Decider v1.1 27B & 72.3 {[}69.5, 74.1{]} & 92.9 {[}92.1, 93.6{]} & 77.1 {[}76.1, 78.2{]} & 9.3 \\
\mdourrow SL-27B Instant & 79.6 {[}78.0, 80.9{]} & 97.4 {[}96.8, 97.9{]} & 75.3 {[}74.3, 76.4{]} & 5.8 \\
\mdourrow SL-27B AutoThink & 93.8 {[}81.9, 96.4{]} & 96.9 {[}96.3, 97.4{]} & 58.3 {[}57.2, 59.5{]} & 6.6 \\
\mdbodyrow Kev 27B & 35.2 {[}33.2, 38.4{]} & 93.8 {[}93.0, 94.7{]} & 34.1 {[}33.0, 35.5{]} & 9.2 \\
\mdbodyrow JEV-27B & 34.8 {[}31.9, 37.6{]} & 88.9 {[}87.9, 89.8{]} & 47.4 {[}46.2, 48.8{]} & 13.0 \\
\mdbodyrow Clef 27B & 14.5 {[}13.4, 17.1{]} & 70.1 {[}68.7, 71.4{]} & 40.8 {[}39.6, 42.1{]} & 22.2 \\
\mdbodyrow Clef flash 9.5B & 3.0 {[}2.4, 4.1{]} & 58.2 {[}57.1, 59.2{]} & 33.2 {[}32.2, 34.2{]} & 24.0 \\
\bottomrule
\end{tabular}
\end{adjustbox}\par}
\end{table}

\begin{table}[!htbp]
\renewcommand{\thetable}{D2}
\centering
\caption{\textbf{Core pair accuracy by task family for selected models.} GLM-5.3's lowest-effort setting includes reasoning. The selected models exhibit different strengths across families, including cases where AutoThink scores below Instant. AF \texttt{answerable\_\allowbreak{}facts}, CC \texttt{clause\_\allowbreak{}change}, DR \texttt{dep\_resolve}, FM \texttt{filter\_\allowbreak{}match}, LC \texttt{log\_\allowbreak{}count\_\allowbreak{}check}, LR \texttt{ledger\_\allowbreak{}reconcile}, PC \texttt{policy\_case}, PI \texttt{perm\_\allowbreak{}inherit}, RR \texttt{record\_rule}, SC \texttt{sched\_\allowbreak{}conflict}, SG \texttt{state\_goal}, TO \texttt{triage\_\allowbreak{}order}, TV \texttt{toolcall\_\allowbreak{}valid}. Hosted-model aggregates appear in Table~\ref{tab:4}.}\label{tab:D2}
{\footnotesize\setlength{\tabcolsep}{4pt}\renewcommand{\arraystretch}{1.1}
\noindent\begin{adjustbox}{max width=\linewidth}
\begin{tabular}{lrrrrrrrrrrrrr}
\toprule
\textbf{system} & \textbf{AF} & \textbf{CC} & \textbf{DR} & \textbf{FM} & \textbf{LC} & \textbf{LR} & \textbf{PC} & \textbf{PI} & \textbf{RR} & \textbf{SC} & \textbf{SG} & \textbf{TO} & \textbf{TV} \\
\midrule
\mdbodyrow GLM-5.3 (chat model), low reasoning effort & 92.7 & 93.0 & 91.7 & 96.0 & 74.0 & 69.7 & 81.6 & 69.3 & 78.7 & 72.3 & 78.7 & 74.6 & 97.0 \\
\mdbodyrow Perplexity Decider v1.1 27B & 92.0 & 96.0 & 94.7 & 37.3 & 72.0 & 41.3 & 97.5 & 78.7 & 50.0 & 51.3 & 76.7 & 33.8 & 90.0 \\
\mdourrow SL-27B Instant & 95.3 & 97.5 & 99.7 & 46.0 & 95.3 & 42.8 & 100.0 & 85.0 & 54.7 & 58.3 & 81.3 & 35.8 & 88.7 \\
\mdourrow SL-27B AutoThink & 93.0 & 80.6 & 84.7 & 83.7 & 84.0 & 87.6 & 86.1 & 80.3 & 94.0 & 85.0 & 93.7 & 83.6 & 93.7 \\
\mdbodyrow Kev 27B & 68.3 & 60.7 & 38.7 & 20.7 & 36.7 & 39.3 & 44.3 & 42.3 & 41.3 & 28.7 & 51.0 & 24.4 & 78.7 \\
\mdbodyrow JEV-27B & 72.0 & 59.2 & 41.7 & 26.0 & 31.7 & 36.3 & 35.8 & 34.0 & 34.7 & 36.0 & 56.3 & 24.4 & 86.7 \\
\mdbodyrow Clef 27B & 39.3 & 26.4 & 12.3 & 14.3 & 10.7 & 31.3 & 27.9 & 29.0 & 22.7 & 21.7 & 27.7 & 15.4 & 45.0 \\
\mdbodyrow DeepSeek-V4-Pro (chat model), no reasoning & 31.7 & 0.0 & 13.3 & 18.7 & 1.0 & 28.9 & 9.5 & 13.3 & 15.3 & 2.7 & 12.3 & 6.5 & 33.3 \\
\mdbodyrow Clef flash 9.5B & 23.7 & 9.0 & 6.0 & 3.7 & 0.0 & 13.4 & 19.9 & 21.3 & 0.7 & 5.0 & 11.3 & 8.5 & 40.7 \\
\bottomrule
\end{tabular}
\end{adjustbox}\par}
\end{table}

\begin{table}[!htbp]
\renewcommand{\thetable}{D3}
\centering
\caption{\textbf{Reference scores by layer.} Task scores are chance-corrected and floored at zero before layer averaging (0 = chance, 100 = perfect). The total averages the seven layers. For the hosted model, the two reranking results use the source authors' published rankings rather than our rerun. The final row reports paired Perplexity-minus-Instant differences with 95\% intervals from 600 bootstrap resamples.}\label{tab:D3}
% Internal analysis provenance (not rendered): Source: \texttt{v1\_\allowbreak{}3/\allowbreak{}derived/\allowbreak{}paired\_\allowbreak{}pplx\_\allowbreak{}minus\_\allowbreak{}27b\_\allowbreak{}instant\_\allowbreak{}v13.txt}.
{\footnotesize\setlength{\tabcolsep}{5pt}\renewcommand{\arraystretch}{1.1}
\noindent\begin{adjustbox}{max width=\linewidth}
\begin{tabular}{>{\raggedright\arraybackslash}p{\mdw{0.1762}}>{\raggedleft\arraybackslash}p{\mdw{0.1015}}>{\raggedleft\arraybackslash}p{\mdw{0.1015}}>{\raggedleft\arraybackslash}p{\mdw{0.1015}}>{\raggedleft\arraybackslash}p{\mdw{0.1015}}>{\raggedleft\arraybackslash}p{\mdw{0.1015}}>{\raggedleft\arraybackslash}p{\mdw{0.1015}}>{\raggedleft\arraybackslash}p{\mdw{0.1015}}>{\raggedleft\arraybackslash}p{\mdw{0.1015}}}
\toprule
\textbf{system} & \textbf{A} & \textbf{B} & \textbf{C} & \textbf{D} & \textbf{E} & \textbf{F} & \textbf{G} & \textbf{total} \\
\midrule
\mdourrow SL-27B Instant & 40.6 & 79.6 & 62.5 & 76.7 & 53.9 & 25.1 & 77.2 & 59.4 \\
\mdourrow SL-27B AutoThink & 40.4 & 80.2 & 62.4 & 74.9 & 54.4 & 25.7 & 74.6 & 58.9 \\
\mdbodyrow Perplexity Decider v1.1 27B & 35.6 & 77.1 & 61.7 & 73.6 & 50.6 & 21.9 & 78.4 & 57.0 \\
\mdbodyrow hosted reference (jev-1.13.0) & 36.5 & 79.6 & 59.4 & 66.5 & 49.8 & 21.3 & 72.3 & 55.0 \\
\mdbodyrow JEV-27B & 41.7 & 71.4 & 57.2 & 69.7 & 46.2 & 23.1 & 71.6 & 54.4 \\
\mdbodyrow Kev 27B & 39.1 & 70.2 & 52.9 & 69.7 & 48.3 & 24.9 & 73.1 & 54.0 \\
\mdbodyrow Clef 27B & 33.1 & 76.3 & 50.7 & 64.1 & 44.2 & 26.7 & 72.0 & 52.4 \\
\mdbodyrow Clef flash 9.5B & 32.5 & 68.2 & 39.8 & 63.0 & 40.2 & 25.7 & 67.7 & 48.2 \\
\mdbodyrow Perplexity minus SL-27B Instant, paired & $-$5.0 {[}$-$6.5, $-$3.3{]} & $-$2.5 {[}$-$4.0, $-$0.8{]} & $-$0.8 {[}$-$2.3, +0.7{]} & $-$3.0 {[}$-$5.7, $-$0.8{]} & $-$3.3 {[}$-$5.7, $-$0.6{]} & $-$3.2 {[}$-$6.6, +0.0{]} & +1.2 {[}+0.2, +2.2{]} & $-$2.4 {[}$-$3.2, $-$1.6{]} \\
\bottomrule
\end{tabular}
\end{adjustbox}\par}
\end{table}

\begingroup
\renewcommand{\thetable}{D4}
\footnotesize\setlength{\tabcolsep}{4pt}\renewcommand{\arraystretch}{1.1}
\FloatBarrier
\begin{longtable}{llrrrrrr}
\caption{\textbf{Chance-corrected score per reference task} for six systems (0 = chance, 100 = perfect, negative values kept). Skill: the skill score against success-rate labels. Hosted: the hosted reference.}\label{tab:D4}\\
\toprule
\textbf{task} & \textbf{metric} & {\begin{tabular}[b]{@{}r@{}}\textbf{SL-27B}\\\textbf{Instant}\end{tabular}} & \textbf{Perplexity} & \textbf{hosted} & \textbf{JEV-27B} & \textbf{Kev} & {\begin{tabular}[b]{@{}r@{}}\textbf{Clef}\\\textbf{27B}\end{tabular}} \\
\midrule
\endfirsthead
\multicolumn{8}{l}{\textit{Table~\thetable{} (continued)}}\\
\toprule
\textbf{task} & \textbf{metric} & {\begin{tabular}[b]{@{}r@{}}\textbf{SL-27B}\\\textbf{Instant}\end{tabular}} & \textbf{Perplexity} & \textbf{hosted} & \textbf{JEV-27B} & \textbf{Kev} & {\begin{tabular}[b]{@{}r@{}}\textbf{Clef}\\\textbf{27B}\end{tabular}} \\
\midrule
\endhead
\bottomrule
\endfoot
\mdbodyrow A \texttt{ambigqa} & accuracy & 35.2 & 4.0 & 29.3 & 26.5 & 5.7 & 5.5 \\
\mdbodyrow A \texttt{cap\_\allowbreak{}boundary\_\allowbreak{}card} & skill & 12.3 & 9.3 & 4.8 & 23.2 & 21.3 & $-$2.5 \\
\mdbodyrow A \texttt{cap\_\allowbreak{}boundary\_\allowbreak{}record} & skill & 20.8 & 17.4 & 23.8 & 29.7 & 27.8 & 6.2 \\
\mdbodyrow A \texttt{freshqa\_\allowbreak{}premise} & auroc & 91.3 & 91.1 & 88.2 & 85.0 & 84.1 & 84.4 \\
\mdbodyrow A \texttt{known\_\allowbreak{}unknown} & accuracy & 71.8 & 72.4 & 54.9 & 57.6 & 70.2 & 71.2 \\
\mdbodyrow A \texttt{selfaware} & accuracy & 12.4 & 19.2 & 17.9 & 28.2 & 25.8 & 31.3 \\
\mdbodyrow B \texttt{aegis2\_\allowbreak{}prompt} & accuracy & 68.4 & 69.6 & 65.6 & 65.8 & 73.0 & 68.2 \\
\mdbodyrow B \texttt{bipia\_\allowbreak{}injection} & accuracy & 79.5 & 84.1 & 95.5 & 50.0 & 36.4 & 93.2 \\
\mdbodyrow B \texttt{browsesafe} & accuracy & 67.6 & 61.0 & 60.2 & 59.4 & 55.4 & 54.2 \\
\mdbodyrow B \texttt{cosapien\_\allowbreak{}policy} & accuracy & 82.8 & 81.8 & 84.8 & 75.7 & 72.6 & 69.6 \\
\mdbodyrow B \texttt{jailbreak\_\allowbreak{}wild} & accuracy & 64.6 & 48.7 & 56.5 & 57.8 & 60.8 & 58.5 \\
\mdbodyrow B \texttt{jbb\_judge} & auroc & 95.1 & 95.1 & 96.2 & 93.9 & 95.0 & 93.8 \\
\mdbodyrow B \texttt{xstest} & auroc & 99.2 & 99.2 & 98.6 & 97.5 & 97.9 & 96.3 \\
\mdbodyrow C \texttt{contextual\_\allowbreak{}judge} & accuracy & 21.6 & 20.0 & 28.8 & 20.0 & 16.5 & 18.4 \\
\mdbodyrow C \texttt{faithbench} & accuracy & 20.1 & 11.4 & 24.2 & 17.4 & 12.1 & 6.7 \\
\mdbodyrow C \texttt{judgebench} & accuracy & 69.0 & 64.8 & 55.4 & 44.7 & 52.5 & 50.9 \\
\mdbodyrow C \texttt{llmbar} & accuracy & 63.4 & 76.3 & 65.1 & 70.7 & 68.5 & 47.6 \\
\mdbodyrow C \texttt{processbench} & accuracy & 71.2 & 70.9 & 67.0 & 64.1 & 60.9 & 60.0 \\
\mdbodyrow C \texttt{rewardbench} & accuracy & 89.6 & 90.3 & 92.8 & 88.1 & 87.6 & 88.5 \\
\mdbodyrow C \texttt{rewardbench2} & accuracy & 74.1 & 76.4 & 75.0 & 77.3 & 74.7 & 68.8 \\
\mdbodyrow C \texttt{squad2\_\allowbreak{}answerable} & accuracy & 91.0 & 83.2 & 67.2 & 75.5 & 50.4 & 64.6 \\
\mdbodyrow D \texttt{agentreward\_\allowbreak{}success} & accuracy & 55.8 & 56.2 & 38.8 & 55.1 & 56.5 & 46.0 \\
\mdbodyrow D \texttt{bfcl\_\allowbreak{}relevance} & accuracy & 81.9 & 76.4 & 81.9 & 75.0 & 69.4 & 70.8 \\
\mdbodyrow D \texttt{mind2web\_\allowbreak{}element} & accuracy & 92.2 & 88.3 & 78.7 & 78.9 & 83.1 & 75.3 \\
\mdbodyrow E \texttt{cuad\_clause} & accuracy & 75.0 & 64.6 & 61.5 & 63.5 & 63.5 & 63.5 \\
\mdbodyrow E \texttt{folio\_\allowbreak{}validation} & accuracy & 77.8 & 76.4 & 74.1 & 60.8 & 67.5 & 55.7 \\
\mdbodyrow E \texttt{legalbench\_\allowbreak{}mix} & accuracy & 62.7 & 61.3 & 63.5 & 60.6 & 62.3 & 57.7 \\
\mdbodyrow E \texttt{primevul\_\allowbreak{}pair} & pair accuracy & $-$22.1 & $-$24.5 & $-$27.4 & $-$23.2 & $-$29.3 & $-$26.6 \\
\mdbodyrow F \texttt{label\_\allowbreak{}errors} & accuracy & 25.1 & 21.9 & 21.3 & 23.1 & 24.9 & 26.7 \\
\mdbodyrow G \texttt{entity\_\allowbreak{}match\_\allowbreak{}abt\_\allowbreak{}buy} & auroc & 99.3 & 99.3 & 97.9 & 99.1 & 98.8 & 98.8 \\
\mdbodyrow G \texttt{entity\_\allowbreak{}match\_\allowbreak{}amazon\_\allowbreak{}google} & auroc & 93.7 & 94.1 & 92.2 & 92.8 & 92.6 & 91.5 \\
\mdbodyrow G \texttt{nevir} & pair accuracy & 71.7 & 73.5 & 57.0 & 61.3 & 65.0 & 63.9 \\
\mdbodyrow G \texttt{paws} & accuracy & 84.2 & 88.5 & 71.5 & 67.8 & 67.2 & 66.0 \\
\mdbodyrow G \texttt{rerank\_\allowbreak{}nfcorpus} & ndcg@10 & 34.3 & 35.1 & 33.3 & 30.3 & 33.2 & 33.2 \\
\mdbodyrow G \texttt{rerank\_\allowbreak{}scifact} & ndcg@10 & 80.0 & 79.9 & 81.8 & 78.3 & 82.0 & 78.4 \\
\end{longtable}
\endgroup

\begin{table}[!htbp]
\renewcommand{\thetable}{D5}
\centering
\caption{\textbf{Reasoning settings on a 1,296-item core subset.} Sampling preserves complete contrast groups and balanced levels: 100 items per family, or 99 for three-member groups. GLM-5.3 uses reasoning at both settings; Instant and Perplexity use one pass. The chat-model reasoning settings also differ in sampling and output budget. Reasoning tokens are averaged per item; a dash denotes no reasoning trace.}\label{tab:D5}
% Internal analysis provenance (not rendered): Sources: \texttt{v1\_\allowbreak{}1/\allowbreak{}derived/\allowbreak{}sample\_\allowbreak{}comparison.csv} and \texttt{v1\_\allowbreak{}1/\allowbreak{}general\_\allowbreak{}llm/}.
{\small\setlength{\tabcolsep}{6pt}\renewcommand{\arraystretch}{1.1}
\noindent\begin{adjustbox}{max width=\linewidth}
\begin{tabular}{lrrrr}
\toprule
\textbf{system} & \textbf{accuracy} & \textbf{pair accuracy} & {\begin{tabular}[b]{@{}r@{}}\textbf{level}\\\textbf{3}\end{tabular}} & {\begin{tabular}[b]{@{}r@{}}\textbf{reasoning}\\\textbf{tokens}\\\textbf{per item}\end{tabular}} \\
\midrule
\mdbodyrow GLM-5.3 (chat model), reasoning on & 97.2 & 93.5 {[}91.7, 95.3{]} & 88.3 & 265 \\
\mdbodyrow GLM-5.3 (chat model), low reasoning effort & 91.7 & 81.5 {[}78.4, 84.6{]} & 73.1 & 121 \\
\mdbodyrow Perplexity Decider v1.1 27B, one pass & 83.6 & 70.4 {[}66.9, 74.0{]} & 56.7 & -- \\
\mdourrow SL-27B Instant, one pass & 87.1 & 75.6 {[}71.9, 79.2{]} & 65.6 & -- \\
\mdourrow SL-27B AutoThink, reasoning below 0.9 & 93.5 & 85.9 {[}82.9, 88.8{]} & 80.4 & 297 \\
\bottomrule
\end{tabular}
\end{adjustbox}\par}
\end{table}

\clearpage
\begin{table}[!htbp]
\renewcommand{\thetable}{D6}
\centering
\caption{\textbf{Probability ordering, label correctness, and difficulty.} For the selected models, the first five columns cover 2,400 pairs from eight binary families; level-1 and level-3 scores average all 13 families. The best single threshold is selected on the evaluation pairs and provides an optimistic diagnostic bound rather than held-out calibration performance.}\label{tab:D6}
% Internal analysis provenance (not rendered): Sources: \texttt{v1\_\allowbreak{}1/\allowbreak{}analysis/\allowbreak{}out/\allowbreak{}t4\_\allowbreak{}threshold\_\allowbreak{}pooled.csv} and \texttt{t2\_\allowbreak{}levels\_\allowbreak{}long.csv}.
{\footnotesize\setlength{\tabcolsep}{6pt}\renewcommand{\arraystretch}{1.1}
\noindent\begin{adjustbox}{max width=\linewidth}
\begin{tabular}{lrrrrrrr}
\toprule
\textbf{system} & \textbf{direction} & {\begin{tabular}[b]{@{}r@{}}\textbf{pair}\\\textbf{accuracy}\end{tabular}} & {\begin{tabular}[b]{@{}r@{}}\textbf{ranked}\\\textbf{right, not}\\\textbf{both right}\end{tabular}} & {\begin{tabular}[b]{@{}r@{}}\textbf{yes member}\\\textbf{answered no}\end{tabular}} & {\begin{tabular}[b]{@{}r@{}}\textbf{pair accuracy,}\\\textbf{best single}\\\textbf{threshold}\end{tabular}} & {\begin{tabular}[b]{@{}r@{}}\textbf{level}\\\textbf{1}\end{tabular}} & {\begin{tabular}[b]{@{}r@{}}\textbf{level}\\\textbf{3}\end{tabular}} \\
\midrule
\mdbodyrow Perplexity Decider v1.1 27B & 92.3 & 74.1 & 18.2 & 7.9 & 74.2 & 87.7 & 57.4 \\
\mdourrow SL-27B Instant & 96.2 & 81.2 & 15.0 & 6.8 & 82.3 & 90.7 & 65.3 \\
\mdourrow SL-27B AutoThink & 97.5 & 87.2 & 10.2 & 5.3 & 87.8 & 89.2 & 82.2 \\
\mdbodyrow Kev 27B & 84.0 & 45.6 & 38.4 & 26.8 & 47.5 & 66.5 & 26.5 \\
\mdbodyrow JEV-27B & 82.9 & 48.0 & 34.8 & 18.5 & 48.3 & 66.4 & 27.0 \\
\mdbodyrow Clef 27B & 78.6 & 25.0 & 53.6 & 44.8 & 30.6 & 42.3 & 12.8 \\
\bottomrule
\end{tabular}
\end{adjustbox}\par}
\end{table}

\begin{table}[!htbp]
\renewcommand{\thetable}{D7}
\centering
\caption{\textbf{Error distributions for selected decision models.} The first two columns give accuracy on cases annotated as having no planted problem or a planted problem, across the 12 annotated families. Gap has a paired 95\% interval. Kinds below 50 counts variation types with pair accuracy below 50\%, among 46 types meeting the inclusion criteria. The Errors column counts incorrect predictions among all 7,800 core items. The final column reports the percentage of incorrect predictions assigned confidence at least 0.90; its denominator is all incorrect predictions.}\label{tab:D7}
% Internal analysis provenance (not rendered): Sources: \texttt{v1\_\allowbreak{}1/\allowbreak{}analysis/\allowbreak{}out/\allowbreak{}t1\_\allowbreak{}kinds\_\allowbreak{}summary.csv} and \texttt{t5\_\allowbreak{}confidence\_\allowbreak{}long.csv}.
{\footnotesize\setlength{\tabcolsep}{6pt}\renewcommand{\arraystretch}{1.1}
\noindent\begin{adjustbox}{max width=\linewidth}
\begin{tabular}{lrrrrrr}
\toprule
\textbf{system} & {\begin{tabular}[b]{@{}r@{}}\textbf{no}\\\textbf{planted}\\\textbf{problem}\end{tabular}} & {\begin{tabular}[b]{@{}r@{}}\textbf{planted}\\\textbf{problem}\end{tabular}} & \textbf{gap} & {\begin{tabular}[b]{@{}r@{}}\textbf{kinds below}\\\textbf{50 (pair}\\\textbf{accuracy)}\end{tabular}} & \textbf{errors} & {\begin{tabular}[b]{@{}r@{}}\textbf{errors at 0.90}\\\textbf{or more (share}\\\textbf{of errors)}\end{tabular}} \\
\midrule
\mdbodyrow Perplexity Decider v1.1 27B & 80.9 & 80.9 & +0.1 {[}$-$1.8, +2.2{]} & 14 of 46 & 1,294 & 33.2 {[}30.5, 36.3{]} \\
\mdourrow SL-27B Instant & 79.3 & 85.6 & +6.3 {[}+4.4, +8.2{]} & 12 of 46 & 1,064 & 14.6 {[}11.9, 17.7{]} \\
\mdourrow SL-27B AutoThink & 94.0 & 93.9 & $-$0.1 {[}$-$1.2, +1.1{]} & 0 of 46 & 470 & 30.2 {[}25.9, 34.7{]} \\
\mdbodyrow Kev 27B & 56.5 & 77.5 & +20.9 {[}+18.5, +23.4{]} & 31 of 46 & 2,297 & 7.1 {[}6.2, 8.1{]} \\
\mdbodyrow JEV-27B & 63.0 & 71.9 & +8.8 {[}+6.3, +11.4{]} & 32 of 46 & 2,316 & 17.8 {[}16.3, 19.3{]} \\
\mdbodyrow Clef 27B & 32.7 & 76.7 & +43.9 {[}+41.7, +46.1{]} & 44 of 46 & 3,119 & 30.6 {[}29.0, 32.1{]} \\
\mdbodyrow Clef flash 9.5B & 28.6 & 65.3 & +36.7 {[}+34.8, +38.6{]} & 43 of 46 & 3,846 & 28.1 {[}27.0, 29.5{]} \\
\bottomrule
\end{tabular}
\end{adjustbox}\par}
\end{table}

\begin{table}[!htbp]
\renewcommand{\thetable}{D8}
\centering
\caption{\textbf{Input sensitivity on version-1.0 core items.} Systems are Perplexity Decider, Kev 27B, JEV-27B, and SL-27B Instant; Instant uses its own rerun of the original items as the baseline. Pair-accuracy changes are ranges across systems. Intervals include zero for all 4 systems under question paraphrasing; they exclude zero for 1 of 4 with the irrelevant paragraph and 2 of 4 with bare option labels. Results use version-1.0 core items, excluding reordered \texttt{log\_\allowbreak{}count\_\allowbreak{}check} runs.}\label{tab:D8}
% Internal analysis provenance (not rendered): Sources: \texttt{v1/\allowbreak{}sens/\allowbreak{}results/} and \texttt{v1\_\allowbreak{}3/\allowbreak{}sens\_\allowbreak{}soup27/\allowbreak{}results/}.
{\footnotesize\setlength{\tabcolsep}{5pt}\renewcommand{\arraystretch}{1.1}
\noindent\begin{adjustbox}{max width=\linewidth}
\begin{tabular}{>{\raggedright\arraybackslash}p{\mdw{0.1591}}>{\raggedright\arraybackslash}p{\mdw{0.4133}}>{\raggedright\arraybackslash}p{\mdw{0.1192}}>{\raggedleft\arraybackslash}p{\mdw{0.1543}}>{\raggedleft\arraybackslash}p{\mdw{0.1407}}}
\toprule
\textbf{variant} & \textbf{what changes} & \textbf{what it tests} & \textbf{pair accuracy change} & \textbf{top option changes} \\
\midrule
\mdbodyrow option order (two files) & the order of the options & position bias & $-$0.5 to 0.0 & 1.5\% to 4.9\% \\
\mdbodyrow paraphrased question & the instruction, one paraphrase per family & wording & +0.1 to +0.6 & 1.5\% to 4.7\% \\
\mdbodyrow irrelevant paragraph & three sentences without facts added to the state & distraction & $-$1.2 to +0.3 & 1.8\% to 6.6\% \\
\mdbodyrow bare option labels & option texts cut to labels such as ``Yes'' (12 families) & option texts & $-$1.7 to $-$0.4 & 1.7\% to 6.3\% \\
\mdbodyrow natural text & states of 8 families rewritten by an LLM as e-mails, tickets or memos, every fact checked & form & $-$0.2 to +1.2 & 5.8\% to 13.9\% \\
\mdbodyrow repeat run & none & run-to-run variability & 0.0 & 0.0\% \\
\bottomrule
\end{tabular}
\end{adjustbox}\par}
\end{table}

\begin{table}[!htbp]
\renewcommand{\thetable}{D9}
\centering
\caption{\textbf{Option-order sensitivity for four models.} Two option-order variants of version-1.0 core items cover 12 families. Pair-accuracy changes have 95\% intervals from 300 group bootstrap resamples. Flips reports the percentage of items whose predicted option changes; the final column separates correct-to-incorrect and incorrect-to-correct changes under the first option-order variant. Intervals containing zero do not demonstrate equivalence.}\label{tab:D9}
% Internal analysis provenance (not rendered): Source: \texttt{v1/\allowbreak{}sens/\allowbreak{}results/\allowbreak{}sens\_\allowbreak{}table.md}.
{\small\setlength{\tabcolsep}{6pt}\renewcommand{\arraystretch}{1.1}
\noindent\begin{adjustbox}{max width=\linewidth}
\begin{tabular}{lrrrrr}
\toprule
\textbf{system} & {\begin{tabular}[b]{@{}r@{}}\textbf{pair accuracy}\\\textbf{change, order 1}\end{tabular}} & {\begin{tabular}[b]{@{}r@{}}\textbf{pair accuracy}\\\textbf{change, order 2}\end{tabular}} & {\begin{tabular}[b]{@{}r@{}}\textbf{accuracy}\\\textbf{change}\\\textbf{(1 / 2)}\end{tabular}} & {\begin{tabular}[b]{@{}r@{}}\textbf{flips (1}\\\textbf{/ 2)}\end{tabular}} & {\begin{tabular}[b]{@{}r@{}}\textbf{right to wrong}\\\textbf{/ wrong to}\\\textbf{right (1)}\end{tabular}} \\
\midrule
\mdbodyrow Perplexity Decider v1.1 27B & 0.0 {[}$-$0.6, +0.5{]} & $-$0.3 {[}$-$1.0, +0.4{]} & $-$0.1 / 0.0 & 2.3 / 2.3 & 0.9 / 0.8 \\
\mdourrow SL-27B Instant & 0.0 {[}$-$0.5, +0.6{]} & 0.0 {[}$-$0.6, +0.7{]} & $-$0.1 / $-$0.1 & 1.5 / 1.9 & 0.7 / 0.5 \\
\mdbodyrow Kev 27B & $-$0.5 {[}$-$1.5, +0.6{]} & $-$0.4 {[}$-$1.4, +0.5{]} & $-$0.5 / $-$0.4 & 4.9 / 4.6 & 2.0 / 1.5 \\
\mdbodyrow JEV-27B & $-$0.1 {[}$-$1.1, +0.8{]} & $-$0.4 {[}$-$1.4, +0.4{]} & +0.1 / $-$0.2 & 4.9 / 4.8 & 1.8 / 1.9 \\
\bottomrule
\end{tabular}
\end{adjustbox}\par}
\end{table}

\begin{table}[!htbp]
\renewcommand{\thetable}{D10}
\centering
\caption{\textbf{General models as reference points: in one pass, the open general models solve 1.5\% to 37.2\% of the contrast groups.} Same columns as Table~\ref{tab:4}. API chat models are evaluated on the core track only; confidence-based coverage is omitted because their probabilities are verbally stated. The open models are read out in one pass from the probabilities of the option keys. $^{\S}$Phi-4 assigns a median of 0.30 of its probability mass to the exact keys, so its scores are based on a renormalized share. $^{\P}$Gemma-4-31B-it assigns less than half of its probability mass to the exact keys on 12\% of the items. $^{\ddagger}$Gemma-3-12B-it is the Unsloth upload of the weights.}\label{tab:Dgen}
{\small\setlength{\tabcolsep}{5pt}\renewcommand{\arraystretch}{1.1}
\noindent\begin{adjustbox}{max width=\linewidth}
\begin{tabular}{lllrrrrr}
\toprule
\textbf{model} & \textbf{provider} & \textbf{size} & \textbf{accuracy} & \textbf{pair accuracy} & {\begin{tabular}[b]{@{}r@{}}\textbf{level}\\\textbf{3}\end{tabular}} & {\begin{tabular}[b]{@{}r@{}}\textbf{auto}\\\textbf{at 5\%}\end{tabular}} & \textbf{reference} \\
\midrule
\mdgrouprow \multicolumn{8}{l}{\textbf{General chat models through an API, stated probabilities}} \\
\mdbodyrow GLM-5.3, lowest reasoning effort & Z.ai & -- & 91.7 & 82.2 {[}80.8, 83.7{]} & 75.6 & -- & -- \\
\mdbodyrow DeepSeek-V4-Pro, no reasoning & DeepSeek & -- & 52.7 & 14.3 {[}13.2, 15.5{]} & 5.7 & -- & -- \\
\mdbodyrow DeepSeek-V4.1-Flash, no reasoning & DeepSeek & -- & 53.1 & 14.2 {[}13.0, 15.5{]} & 4.4 & -- & -- \\
\midrule
\mdgrouprow \multicolumn{8}{l}{\textbf{Open general models, one pass, probabilities of the option keys}} \\
\mdbodyrow Qwen3.8-27B & Qwen & 27B & 53.6 & 12.9 {[}11.6, 14.2{]} & 3.2 & 12.1 & 51.6 \\
\mdbodyrow Qwen3.6-35B-A3B & Qwen & 35B-A3B & 49.1 & 9.9 {[}8.9, 10.9{]} & 4.0 & 6.4 & 49.9 \\
\mdbodyrow Gemma-4-31B-it$^{\P}$ & Google & 31B & 67.5 & 37.2 {[}35.5, 39.0{]} & 15.8 & 22.3 & 54.4 \\
\mdbodyrow Gemma-4-26B-A4B-it & Google & 26B-A4B & 56.6 & 20.4 {[}18.9, 21.7{]} & 7.9 & 10.9 & 51.6 \\
\mdbodyrow Phi-4$^{\S}$ & Microsoft & 14B & 46.5 & 8.2 {[}7.4, 9.2{]} & 4.1 & 0.9 & 42.9 \\
\mdbodyrow Gemma-3-12B-it$^{\ddagger}$ & Google & 12B & 44.7 & 8.5 {[}7.7, 9.3{]} & 6.5 & 1.0 & 40.2 \\
\mdbodyrow Qwen3.5-9B & Qwen & 9B & 47.9 & 11.4 {[}10.3, 12.4{]} & 7.6 & 2.9 & 47.8 \\
\mdbodyrow Granite-3.3-8B-Instruct & IBM & 8B & 42.0 & 4.8 {[}4.2, 5.5{]} & 3.3 & 0.3 & 38.3 \\
\mdbodyrow Qwen3.5-4B & Qwen & 4B & 46.6 & 9.6 {[}8.8, 10.4{]} & 6.2 & 2.2 & 41.7 \\
\mdbodyrow Qwen3.5-2B & Qwen & 2B & 40.8 & 3.6 {[}3.1, 4.2{]} & 3.5 & 0.7 & 25.9 \\
\mdbodyrow Qwen3.5-0.8B & Qwen & 0.8B & 40.0 & 1.5 {[}1.1, 1.8{]} & 1.7 & 0.1 & 11.0 \\
\bottomrule
\end{tabular}
\end{adjustbox}\par}
\end{table}

\FloatBarrier
\begin{figure}[!htbp]
\renewcommand{\thefigure}{D1}
\centering
\adjincludegraphics[max width=\linewidth]{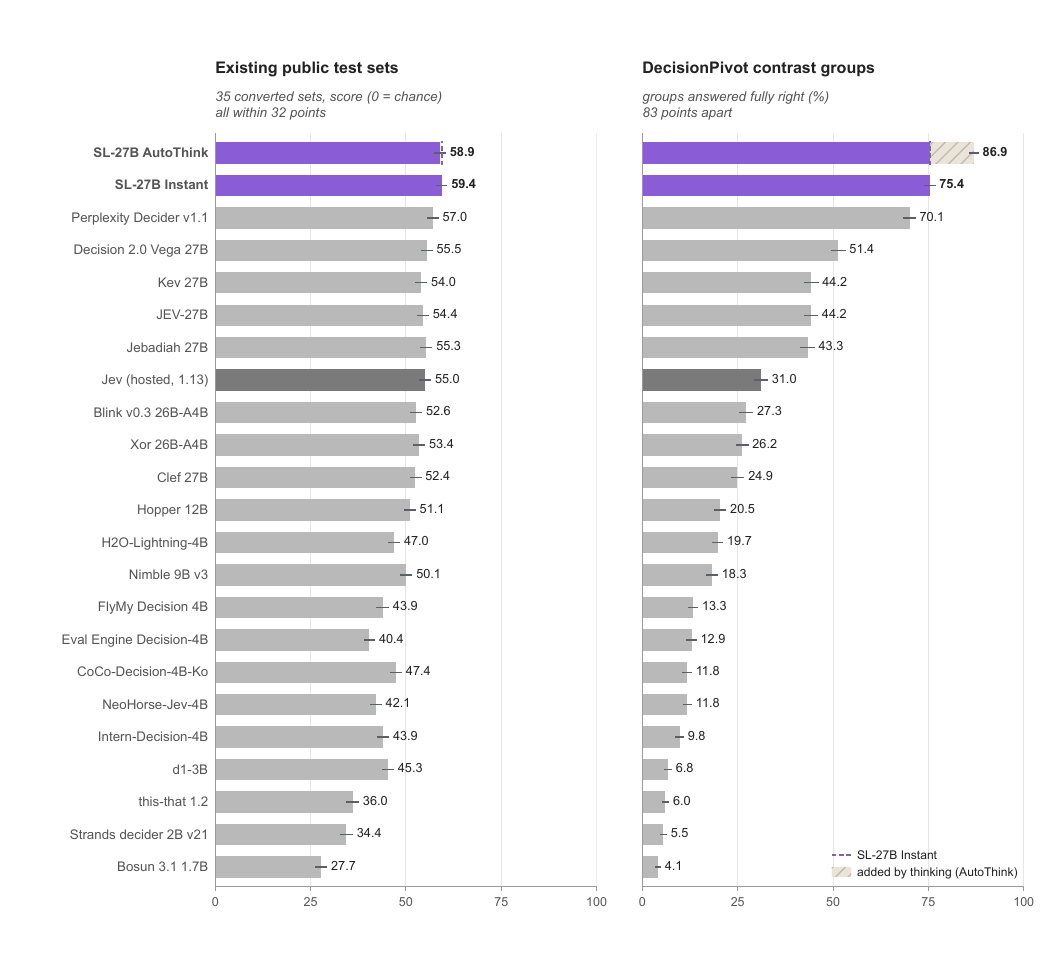}
\caption{\textbf{Reference and core scores for a selected model subset.} One selected model per organization, plus both SL-27B variants. Left: aggregate reference score (0 = chance). Right: core pair accuracy. Models appear in the same order with 95\% intervals. The plotted models span approximately 32 reference-score points and 83 percentage points in core pair accuracy; the tracks differ in task content and scoring. The full main comparison is in Table~\ref{tab:4}.}\label{fig:D2}
\end{figure}

\begin{figure}[!htbp]
\renewcommand{\thefigure}{D2}
\centering
\adjincludegraphics[max width=\linewidth]{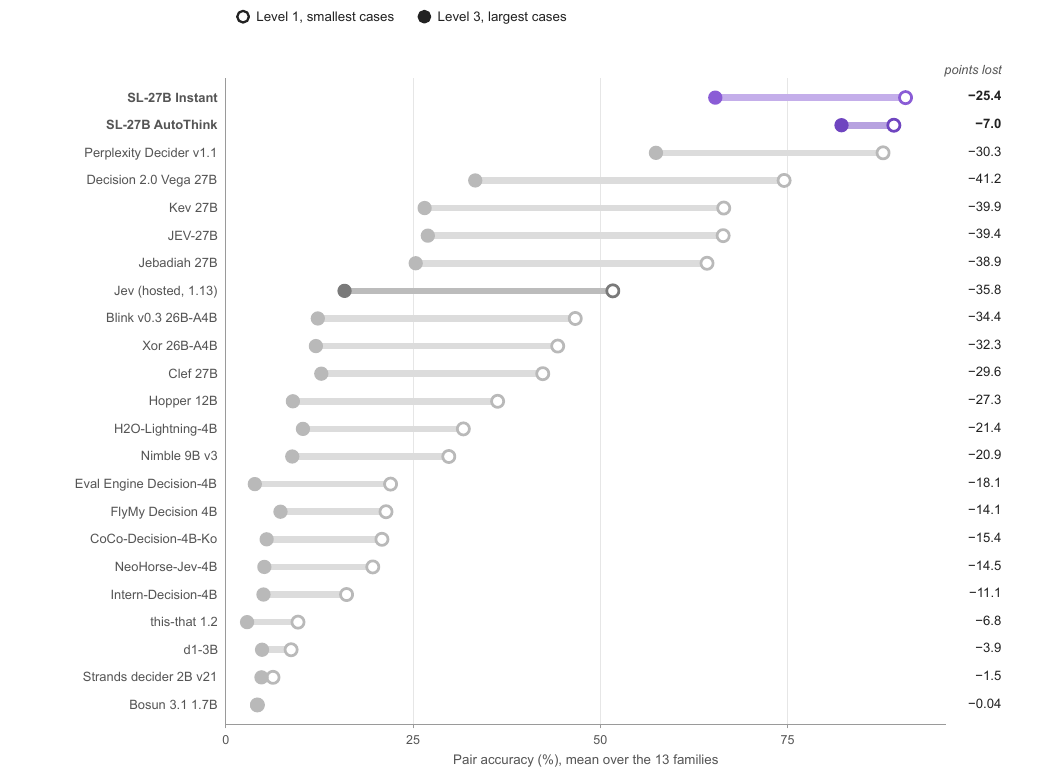}
\caption{\textbf{Performance declines with increased task difficulty.} Level-1 and level-3 pair accuracy, averaged over 13 families, for the selected models shown. Open markers denote level 1 and filled markers level 3; labels give the change in percentage points. Higher levels jointly increase case size and complexity. For example, \texttt{sched\_conflict} grows from 2 rooms and 6--8 bookings to 5--6 rooms and 20--28 bookings.}\label{fig:D1}
\end{figure}

\FloatBarrier
\section{Audit of the generators}\label{sec:E}

\begin{table}[!htbp]
\renewcommand{\thetable}{E1}
\centering
\caption{\textbf{Generator audit results for version 1.1.} Ref denotes the maximum of the majority-class rate, the most frequent gold-position share, and uniform-guess accuracy. Options, question + options, and state words are cross-validated bag-of-words probes. Length and lines use the best single threshold for yes/no tasks. These probes assess specific shortcuts rather than all possible heuristics. Solver reports agreement with gold answers; a second entry indicates an additional draw for each of the three unchanged families.}\label{tab:E1}
% Internal analysis provenance (not rendered): Sources: \texttt{gen\_\allowbreak{}v4/\allowbreak{}audit\_\allowbreak{}v3.py}, \texttt{solve\_\allowbreak{}all.py}, and \texttt{AUDIT.md}.
{\small\setlength{\tabcolsep}{6pt}\renewcommand{\arraystretch}{1.1}
\noindent\begin{adjustbox}{max width=\linewidth}
\begin{tabular}{lrrrrrrrr}
\toprule
\textbf{family} & \textbf{ref} & \textbf{options} & {\begin{tabular}[b]{@{}r@{}}\textbf{question}\\\textbf{+}\\\textbf{options}\end{tabular}} & \textbf{length} & \textbf{lines} & {\begin{tabular}[b]{@{}r@{}}\textbf{state}\\\textbf{words}\end{tabular}} & \textbf{domains} & \textbf{solver} \\
\midrule
\mdbodyrow \texttt{perm\_\allowbreak{}inherit} & 0.500 & 0.472 & 0.472 & 0.503 & 0.500 & 0.500 & 5 & 600 / 600, 600 / 600 \\
\mdbodyrow \texttt{sched\_\allowbreak{}conflict} & 0.500 & 0.464 & 0.464 & 0.502 & 0.500 & 0.527 & 5 & 600 / 600 \\
\mdbodyrow \texttt{triage\_\allowbreak{}order} & 0.200 & 0.213 & 0.213 & -- & -- & 0.193 & 6 & 600 / 600 \\
\mdbodyrow \texttt{clause\_\allowbreak{}change} & 0.333 & 0.253 & 0.253 & -- & -- & 0.305 & 6 & 600 / 600, 600 / 600 \\
\mdbodyrow \texttt{dep\_resolve} & 0.500 & 0.469 & 0.469 & 0.503 & 0.500 & 0.500 & 5 & 600 / 600, 600 / 600 \\
\mdbodyrow \texttt{ledger\_\allowbreak{}reconcile} & 0.167 & 0.146 & 0.144 & -- & -- & 0.128 & 5 & 600 / 600 \\
\mdbodyrow \texttt{record\_rule} & 0.122 & 0.111 & 0.110 & -- & -- & 0.116 & 8 & 600 / 600 \\
\mdbodyrow \texttt{answerable\_\allowbreak{}facts} & 0.500 & 0.469 & 0.469 & 0.507 & 0.500 & 0.489 & -- & 600 / 600 \\
\mdbodyrow \texttt{policy\_case} & 0.333 & 0.260 & 0.260 & -- & -- & 0.238 & 4 & 600 / 600 \\
\mdbodyrow \texttt{log\_\allowbreak{}count\_\allowbreak{}check} & 0.500 & 0.460 & 0.460 & 0.503 & 0.500 & 0.457 & -- & 600 / 600 \\
\mdbodyrow \texttt{state\_goal} & 0.500 & 0.500 & 0.500 & 0.505 & 0.500 & 0.500 & 5 & 600 / 600 \\
\mdbodyrow \texttt{toolcall\_\allowbreak{}valid} & 0.500 & 0.500 & 0.500 & 0.507 & 0.508 & 0.527 & -- & 600 / 600 \\
\mdbodyrow \texttt{filter\_\allowbreak{}match} & 0.500 & 0.473 & 0.473 & 0.503 & 0.500 & 0.511 & 4 & 600 / 600 \\
\bottomrule
\end{tabular}
\end{adjustbox}\par}
\end{table}

The \href{https://github.com/StartLuxLabs/DecisionPivot/tree/2b7000f312ccbbdce4e8625b56d8a8186ea0047a/audits}{audit implementation} and \href{https://github.com/StartLuxLabs/DecisionPivot/blob/2b7000f312ccbbdce4e8625b56d8a8186ea0047a/docs/AUDIT.md}{supporting audit report} are available in the benchmark code repository.
No state is repeated. Across the binary and three-way families, no word appearing at least 20 times is more than twice as frequent in one class as in another.

\textbf{AI-assisted item audit.} Two AI reviewers independently answered 65 version-1.0 items without access to the gold labels and their answers agreed with all 65 gold labels. They flagged two \texttt{filter\_\allowbreak{}match} items as potentially ambiguous and rated 53--55 items as only partly natural.

\textbf{Changes in version 1.1.} The identified issues involve ambiguous or artificial wording and exploitable shortcuts rather than changes to the intended gold answers. Ten families were revised accordingly (Table~\ref{tab:E2}). Partial-information heuristics can still succeed: in \texttt{policy\_case}, selecting the least risky-looking case identifies the approve case in 44\% of triples, compared with 33\% chance; in \texttt{toolcall\_valid}, checking the schema alone achieves 0.917 item accuracy. These findings qualify the shortcut-audit results.

\textbf{Rank stability.} Across 20 configurations evaluated on core versions 1.0 and 1.1, the Spearman rank correlation is 0.986, with pair-accuracy changes ranging from $-$2.3 to +4.2 percentage points. This comparison includes checkpoints outside the main evaluation roster. Two subsets of this comparison, comprising seven external configurations and eleven StartLux checkpoints, have median gains of 1.3 and 0.3 points, respectively. On a fresh draw of 4,200 items (Table~\ref{tab:E3}), pair-accuracy changes across 34 decision models range from $-$2.0 to +1.9 points, and 550 of 561 model pairs retain their relative ordering. Changes for the eleven open general models range from $-$0.5 to +0.9 points. These results characterize version and sampling sensitivity within the existing generator designs, rather than ruling out design bias.
% TODO: list all 20 historical configurations and their baseline versions; explain the remaining two configurations beyond the seven external and eleven internal checkpoints.

\clearpage
\begin{table}[!htbp]
\renewcommand{\thetable}{E2}
\centering
\caption{\textbf{Measured item defects and revisions from version 1.0 to 1.1.} Ten families were revised to address wording, consistency, and shortcut issues. These revisions do not establish that all potential ambiguities or shortcuts have been eliminated. Measurements use the same audit procedure on both versions.}\label{tab:E2}
% Internal audit provenance (not rendered): gen_v4/AUDIT.md; public counterpart: docs/AUDIT.md.
{\footnotesize\setlength{\tabcolsep}{5pt}\renewcommand{\arraystretch}{1.1}
\noindent\begin{adjustbox}{max width=\linewidth}
\begin{tabular}{>{\raggedright\arraybackslash}p{\mdw{0.1385}}>{\raggedright\arraybackslash}p{\mdw{0.4531}}>{\raggedright\arraybackslash}p{\mdw{0.3943}}}
\toprule
\textbf{family} & \textbf{defect in version 1.0} & \textbf{change in version 1.1} \\
\midrule
\mdbodyrow \texttt{filter\_\allowbreak{}match} & unspecified day-count conventions, ambiguous field references, and implausible records such as a list price below the price & explicit day-count rules, unambiguous field references, and prices and specifications within their category ranges \\
\mdbodyrow \texttt{state\_goal} & step descriptions inconsistent with some steps in most items, pseudo-word names, and a shortcut in 66 negatives whose failing condition no step touched & consistent step descriptions, ordinary names in five domains, goal-directed traces, the shortcut removed, and pairs annotated as groups \\
\mdbodyrow \texttt{policy\_case} & rules against the direction of risk (approve because consent is missing), dead and near-duplicate rules, inconsistent case records & conditions aligned with rule outcomes, no dead or duplicate rules, and consistent records \\
\mdbodyrow \texttt{record\_rule} & derived fields that disagree outside the planted violation, the status convention of reference lists left implicit & consistent background records, the status convention written in every state that needs it \\
\mdbodyrow \texttt{triage\_\allowbreak{}order} & time columns that contradict each other, a class-first reading that gave the key in 37\% of the level-1 items & consistent times, the class as the first written rule at level 1, the shortcut capped at levels 2 and 3 \\
\mdbodyrow \texttt{log\_\allowbreak{}count\_\allowbreak{}check} & claims such as ``exactly 1 ERROR entries'' and claims open to an at-least reading & explicit exact-count wording, correct noun agreement, and stated counts balanced within each level \\
\mdbodyrow \texttt{sched\_\allowbreak{}conflict} & 60\% of the negatives at levels 2 and 3 resolved by a simple check of hours, seats, or equipment & 30\% at level 2 and 15\% at level 3, remaining violations placed in booking records \\
\mdbodyrow \texttt{answerable\_\allowbreak{}facts} & a quarter of the founding years at one clamped value, two crossed difficulty factors rather than an ordered difficulty progression & founding years spread over a wide range, with three ordered difficulty levels \\
\mdbodyrow \texttt{ledger\_\allowbreak{}reconcile} & a ledger named in two ways within one item, no rule for negative amounts & one name per ledger, the deduction rule stated for both signs \\
\mdbodyrow \texttt{toolcall\_\allowbreak{}valid} & request words equal to a parameter name without referring to it, a few filler sentences repeated across many items & parameter-name cues removed, neutral filler wording diversified, and pairs annotated as groups \\
\bottomrule
\end{tabular}
\end{adjustbox}\par}
\end{table}

\begin{table}[!htbp]
\renewcommand{\thetable}{E3}
\centering
\caption{\textbf{Fresh-sample stability within seven seeded generators.} Pair accuracy (\%) on version-1.1 items and 4,200 fresh items (600 per family) for \texttt{clause\_\allowbreak{}change}, \texttt{dep\_\allowbreak{}resolve}, \texttt{ledger\_\allowbreak{}reconcile}, \texttt{perm\_\allowbreak{}inherit}, \texttt{record\_rule}, \texttt{sched\_\allowbreak{}conflict}, and \texttt{triage\_\allowbreak{}order}. Fresh items were generated with a new seed after the initial evaluations and were not subsequently revised. Change denotes second-draw minus version-1.1 pair accuracy, with 95\% percentile intervals from 2,000 bootstrap resamples of contrast groups within each family, sampled independently for the two sets. Unanswered items count as wrong. The analysis covers a subset of 34 decision models from Table~\ref{tab:4}. Spearman correlation is 0.994, and 550 of 561 model pairs retain their ordering. These results assess sampling sensitivity within the same generator designs.}\label{tab:E3}
{\footnotesize\setlength{\tabcolsep}{5pt}\renewcommand{\arraystretch}{1.1}
\noindent\begin{adjustbox}{max width=\linewidth}
\begin{tabular}{lrrrr}
\toprule
\textbf{decision model} & \textbf{version 1.1} & \textbf{second draw} & \textbf{change} & \textbf{95\% interval} \\
\midrule
\mdourrow \textbf{SL-27B AutoThink} & 85.1 & 84.9 & $-$0.2 & {[}$-$2.8, +2.1{]} \\
\mdourrow \textbf{SL-27B Instant} & 67.7 & 67.9 & +0.3 & {[}$-$2.6, +3.2{]} \\
    \mdbodyrow Perplexity Decider v1.1 & 63.7 & 65.0 & +1.4 & {[}$-$1.5, +4.4{]} \\
\mdbodyrow Decision 2.0 Vega & 42.2 & 44.1 & +1.9 & {[}$-$1.5, +5.1{]} \\
\mdbodyrow Kev 27B & 39.3 & 41.1 & +1.7 & {[}$-$1.5, +5.0{]} \\
\mdbodyrow JEV-27B & 38.0 & 37.8 & $-$0.2 & {[}$-$3.6, +3.0{]} \\
\mdbodyrow Jebadiah 27B & 35.6 & 35.2 & $-$0.4 & {[}$-$3.8, +2.9{]} \\
\mdbodyrow Jev 1.13 (hosted) & 28.8 & 30.2 & +1.4 & {[}$-$1.7, +4.4{]} \\
\mdbodyrow Blink v0.3 & 25.2 & 25.6 & +0.4 & {[}$-$2.7, +3.3{]} \\
\mdbodyrow Clef & 22.7 & 22.9 & +0.3 & {[}$-$2.6, +3.2{]} \\
\mdbodyrow Xor 1.0 & 21.5 & 22.2 & +0.7 & {[}$-$2.3, +3.6{]} \\
\mdbodyrow GEV-26B-Decide & 21.5 & 21.5 & +0.0 & {[}$-$2.7, +2.8{]} \\
\mdbodyrow Hopper 12B & 17.0 & 17.1 & +0.1 & {[}$-$2.4, +2.7{]} \\
\mdbodyrow H2O-Lightning-4B & 15.0 & 15.1 & +0.1 & {[}$-$2.4, +2.4{]} \\
\mdbodyrow JEV-9B & 14.6 & 14.1 & $-$0.4 & {[}$-$2.8, +2.1{]} \\
\mdbodyrow Nimble v3 & 13.2 & 13.0 & $-$0.2 & {[}$-$2.5, +2.0{]} \\
\mdbodyrow Decision 2.0 Lux & 12.2 & 12.1 & $-$0.1 & {[}$-$2.4, +2.1{]} \\
\mdbodyrow Decision 2.0 Nox & 11.0 & 9.0 & $-$2.0 & {[}$-$4.1, $-$0.1{]} \\
\mdbodyrow Eval Engine Decision-4B & 10.4 & 9.7 & $-$0.7 & {[}$-$2.8, +1.2{]} \\
\mdbodyrow FlyMy Decision 4B v1.2 & 10.3 & 10.1 & $-$0.3 & {[}$-$2.1, +1.7{]} \\
\mdbodyrow CoCo-Decision-4B-Ko & 9.9 & 9.0 & $-$1.0 & {[}$-$2.9, +1.0{]} \\
\mdbodyrow Hopper (G) 1.2 & 9.3 & 9.7 & +0.5 & {[}$-$1.4, +2.4{]} \\
\mdbodyrow Clef flash & 9.1 & 8.4 & $-$0.7 & {[}$-$2.4, +1.1{]} \\
\mdbodyrow NeoHorse-Jev-4B & 8.0 & 8.6 & +0.6 & {[}$-$1.2, +2.4{]} \\
\mdbodyrow Intern-Decision-4B & 6.7 & 7.1 & +0.5 & {[}$-$1.2, +2.1{]} \\
\mdbodyrow Strands decider 2B v21 & 4.0 & 4.5 & +0.5 & {[}$-$0.7, +1.7{]} \\
\mdbodyrow d1-3B & 4.0 & 4.6 & +0.7 & {[}$-$0.5, +1.9{]} \\
\mdbodyrow this-that 1.2 & 3.9 & 3.9 & +0.0 & {[}$-$1.2, +1.3{]} \\
\mdbodyrow Decision 2.0 Sol & 3.8 & 3.8 & +0.0 & {[}$-$1.1, +1.2{]} \\
\mdbodyrow d1-omni-600M & 3.4 & 2.8 & $-$0.6 & {[}$-$1.7, +0.4{]} \\
\mdbodyrow Intern-Decision-2B & 3.0 & 3.0 & $-$0.1 & {[}$-$1.1, +1.0{]} \\
\mdbodyrow Decision 2.0 Eos & 3.0 & 2.5 & $-$0.5 & {[}$-$1.5, +0.5{]} \\
\mdbodyrow Bosun 3.1 1.7B & 2.4 & 1.4 & $-$1.1 & {[}$-$1.9, $-$0.3{]} \\
\mdbodyrow Decision 2.0 Kai & 2.0 & 2.2 & +0.2 & {[}$-$0.6, +1.1{]} \\
\bottomrule
\end{tabular}
\end{adjustbox}\par}
\end{table}

\clearpage
\section{Datasheet}\label{sec:F}

Table~\ref{tab:F1} presents the datasheet for the evaluated DecisionPivot release, with version-1.1 core items and version-1.3 reference items, following the framework of \citet{gebru2021datasheets}.
It summarizes the benchmark's construction, composition, evaluation scope, and intended uses, with references to further details in the paper.

\begingroup
\renewcommand{\thetable}{F1}
\footnotesize\setlength{\tabcolsep}{5pt}\renewcommand{\arraystretch}{1.1}
\FloatBarrier
\begin{longtable}{>{\raggedright\arraybackslash}p{\mdw{0.1955}}>{\raggedright\arraybackslash}p{\mdw{0.7902}}}
\caption{\textbf{Datasheet for the evaluated DecisionPivot release.} Questions follow the categories of \citet{gebru2021datasheets}, with answers and references to supporting details in this paper.}\label{tab:F1}\\
\toprule
\textbf{question} & \textbf{answer} \\
\midrule
\endfirsthead
\multicolumn{2}{l}{\textit{Table~\thetable{} (continued)}}\\
\toprule
\textbf{question} & \textbf{answer} \\
\midrule
\endhead
\bottomrule
\endfoot
\mdgrouprow \multicolumn{2}{l}{\textbf{Motivation}} \\
\mdbodyrow Why was the benchmark created? & To diagnose decision correctness, sensitivity to decisive changes, and confidence reliability across seven decision roles in LLM systems. Public evaluations provide useful comparisons but do not alone establish consistent performance across controlled contrasts (Section~\ref{sec:1}). \\
\mdbodyrow Who created it? & StartLux Labs, which also develops the StartLux-Decision models (Section~\ref{sec:6}). \\
\midrule
\mdgrouprow \multicolumn{2}{l}{\textbf{Composition}} \\
\mdbodyrow What does an item represent? & A state, a typed question, an option set, and an answer label. Core items have a gold option and record their contrast group, difficulty level, and, where available, variation type. Reference tasks may also use success-rate labels or ranking judgments (Section~\ref{sec:3.1}). \\
\mdbodyrow How many items are there? & 40,033 items across 48 tasks: 7,800 items from 13 generated core families (600 per family) and 32,233 items from 35 public reference test sets. \\
\mdbodyrow Are the items a sample of a larger set? & Core items are generated, with 200 items per family at each difficulty level. Seven generators support fresh draws with configurable seeds. Reference tasks use fixed-seed sampling with task-specific limits, and final counts reflect filtering (Sections~\ref{sec:3.2.1} and \ref{sec3.2.2}). \\
\mdbodyrow What does an item contain? & Written content comprising a state in text or structured form, a question, an option set, and an answer label. Core items additionally include the annotations described above. Median core state lengths range from 1,647 to 4,975 characters across families (Table~\ref{tab:2}). \\
\mdbodyrow Is there a label? & Yes. Core answers are computed from structured cases, with independent solvers agreeing with all 600 gold answers in each family. Reference labels are human-annotated for 21 tasks, programmatically computed for six, and derived from mixed sources for eight (Table~\ref{tab:C1}). \\
\mdbodyrow Are there recommended splits? & No. All items are intended for evaluation, with no training split. \\
\mdbodyrow Are there errors, noise or redundancies? & Residual errors in reference labels have not been quantified. Content-blind baselines exceed chance on some reference tasks (Section~\ref{sec:quality}). No core state is repeated (Appendix~\ref{sec:E}). \\
\mdbodyrow Is it self-contained? & The release combines generated core items, converted reference data, and reconstruction support. Scripts for 13 reference tasks reconstruct inputs from official source files; source-specific terms and release routes are documented separately (Tables~\ref{tab:C2} and \ref{tab:C4}). \\
\mdbodyrow Does it hold personal, confidential or offensive data? & Core items use invented entities and contain no personal data. Layer B includes harmful reference text as classification inputs (Section~\ref{sec:license}). \\
\midrule
\mdgrouprow \multicolumn{2}{l}{\textbf{Collection process}} \\
\mdbodyrow How were the items made? & Python generators compute core answers from structured cases and render related cases as contrast groups. Most groups vary a decisive detail, while \texttt{policy\_case} groups different cases under a shared policy. Reference items are reconstructed from official source files (Sections~\ref{sec:3.2.1} and \ref{sec3.2.2}). \\
\mdbodyrow Who was involved? & The authors developed the generators, and for each family, an author unfamiliar with the generator independently implemented a solver. All 7,800 final core items were independently screened by at least two human reviewers, with additional AI-assisted screening. A separate AI audit examined 65 version-1.0 items (Appendix~\ref{sec:E}). \\
\mdbodyrow Over what time frame? & Core version 1.1 is dated 8 October 2026; the filtered reference version 1.3 is dated 10 October 2026. Revisions and subsequent fresh-sample evaluations are described in Section~\ref{sec:quality}. \\
\midrule
\mdgrouprow \multicolumn{2}{l}{\textbf{Preprocessing}} \\
\mdbodyrow What preprocessing was done? & We removed 3,016 reference items (8.5\%) detected in the training files of our final 27B models, together with affected groups. Each task was converted into a typed decision interface with a fixed option order, preserving its original evaluation metric and recording source-exposure flags (Sections~\ref{sec3.2.2} and \ref{sec:3.4}). Version 1.1 also revised ten core families to address identified defects (Table~\ref{tab:E2}). \\
\mdbodyrow Is the raw data kept? & Original reference data remain available through their official sources. Reconstruction scripts are provided for 13 tasks, with counts and hashes checked against the evaluated files (Section~\ref{sec:license}). \\
\mdbodyrow Is the software available? & The release includes generators, independent solvers, audit scripts, and scoring code. \\
\midrule
\mdgrouprow \multicolumn{2}{l}{\textbf{Uses}} \\*
\mdbodyrow What has it been used for? & Evaluating 40 decision models and 14 general models in this study (Section~\ref{sec:4}). \\
\mdbodyrow What else could it be used for? & Comparing decision and general models under documented probability-readout protocols, and analyzing contrast-group correctness, confidence reliability, decision errors, and empirical selective coverage (Sections~\ref{sec:3.4} and \ref{sec:5}). \\
\mdbodyrow What could limit future uses? & Core cases are generated rather than sampled from live traffic. Coverage varies across decision roles, and probability interfaces differ across models. Empirical selective coverage does not guarantee deployment performance. Multilingual evaluation, standardized inference costs, and end-to-end system effects remain outside the current scope (Section~\ref{sec:6}). \\
\mdbodyrow What should it not be used for? & The stated access conditions prohibit using evaluation items for model training (Section~\ref{sec:license}). \\
\midrule
\mdgrouprow \multicolumn{2}{l}{\textbf{Distribution}} \\
\mdbodyrow How is it distributed? & The release combines generated core items, converted reference data, reconstruction scripts, generators, solvers, audits, scoring code, and evaluation records. Code: \url{https://github.com/StartLuxLabs/DecisionPivot}; data access is granted on request at \url{https://huggingface.co/datasets/startlux-models/DecisionPivot}. Source terms and reconstruction routes are documented in Table~\ref{tab:C4}. \\
\mdbodyrow Under which license? & Core items and documentation are released under CC BY 4.0, and code under Apache-2.0. Reference items retain their original copyrights and source-specific licenses, as documented in Tables~\ref{tab:license} and \ref{tab:C4}. \\
\mdbodyrow Do third parties restrict its use? & Yes. Reference items remain subject to their original source terms, including non-commercial and research-use restrictions where applicable (Table~\ref{tab:C4}). \\
\midrule
\mdgrouprow \multicolumn{2}{l}{\textbf{Maintenance}} \\
\mdbodyrow Who maintains it, and how can they be reached? & StartLux Labs. Questions and issue reports can be submitted through the benchmark's \href{https://github.com/StartLuxLabs/DecisionPivot/issues}{GitHub issue tracker}. \\
\mdbodyrow Will it be updated? & Future updates may include additional models and fresh evaluation samples from the seven generators with configurable seeds. Each core family is identified by the hash of its items file, and the scorer validates this hash against the evaluated files (Section~\ref{sec:3.4}). \\
\end{longtable}
\endgroup

\mdBibliography
\end{document}